\def\year{2020}\relax
\documentclass[letterpaper]{article} 
\usepackage{aaai20}  
\usepackage{times}  
\usepackage{helvet} 
\usepackage{courier}  
\usepackage[hyphens]{url}  
\usepackage{graphicx} 
\usepackage{graphicx}  
\usepackage{verbatim}
\usepackage{subcaption}
\usepackage{amsthm}
\usepackage{amsmath} 
\usepackage{amsfonts}
\usepackage{bm} 

\title{Neighborhood Cognition Consistent Multi-Agent Reinforcement Learning}
\author{Hangyu Mao,\textsuperscript{\rm 1,2}
	Wulong Liu,\textsuperscript{\rm 2}
	Jianye Hao,\textsuperscript{\rm 3,2}
	Jun Luo\textsuperscript{\rm 2}\\
	\bf \Large Dong Li,\textsuperscript{\rm 2}
	Zhengchao Zhang,\textsuperscript{\rm 1}
	Jun Wang,\textsuperscript{\rm 4}
	Zhen Xiao\textsuperscript{\rm 1}\\
	\textsuperscript{\rm 1}Peking University,
	\textsuperscript{\rm 2}Noah's Ark Lab, Huawei\\
	\textsuperscript{\rm 3}Tianjin University,
	\textsuperscript{\rm 4}University College London\\
	\{hy.mao, zhengchaozhang, xiaozhen\}@pku.edu.cn\\
	\{liuwulong, haojianye, jun.luo1, lidong106\}@huawei.com\\
	jun.wang@cs.ucl.ac.uk
}

\begin{document}

\maketitle

\begin{abstract}
Social psychology and real experiences show that cognitive consistency plays an important role to keep human society in order: if people have a more consistent cognition about their environments, they are more likely to achieve better cooperation. Meanwhile, only cognitive consistency within a neighborhood matters because humans only interact directly with their neighbors. Inspired by these observations, we take the first step to introduce \emph{neighborhood cognitive consistency} (NCC) into multi-agent reinforcement learning (MARL). Our NCC design is quite general and can be easily combined with existing MARL methods. As examples, we propose neighborhood cognition consistent deep Q-learning and Actor-Critic to facilitate large-scale multi-agent cooperations. Extensive experiments on several challenging tasks (i.e., packet routing, wifi configuration and Google football player control) justify the superior performance of our methods compared with state-of-the-art MARL approaches.
\end{abstract}

\section{Introduction}
In social psychology, cognitive consistency theories show that people usually seek to perceive the environment in a simple and consistent way \cite{simon2004redux,russo2008goal,lakkaraju2019cognitive}. If the perceptions are inconsistent, people produce an uncomfortable feeling (i.e., the cognitive dissonance), and they will change behaviors to reduce this feeling by making their cognitions consistent.

Naturally, this also applies to multi-agent systems \cite{oroojlooyjadid2019review,bear2017co,corgnet2015cognitive}: agents that maintain consistent cognitions about their environments are crucial for achieving effective system-level cooperation. In contrast, it is hard to imagine that the agents without consensuses on their situated environments can cooperate well. Besides, the fact that agents only interact directly with their neighbors with local perception indicates that maintaining neighborhood cognitive consistency is usually sufficient to guarantee system-level cooperations.

Inspired by the above observations, we introduce \emph{neighborhood cognitive consistency} into multi-agent reinforcement learning (MARL) to facilitate agent cooperations. Compared to recent MARL methods that focus on designing global coordination mechanisms like centralized critic \cite{Lowe2017Multi,foerster2018counterfactual,mao2019modelling}, joint value factorization \cite{sunehag2018value,rashid2018qmix,son2019qtran} and agent communication \cite{foerster2016learning,sukhbaatar2016learning,peng2017multiagent}, the neighborhood cognitive consistency takes an alternative but complementary approach of focusing on innovating value network design from a local perspective. The benefit is that it can be easily combined with existing MARL methods to facilitate large-scale agent cooperations. As examples, we propose Neighborhood Cognition Consistent Q-learning and Actor-Critic for practical use.

Specifically, we take three steps to implement the neighborhood cognitive consistency. First, we cast a multi-agent environment as a graph, and adopt graph convolutional network to extract a high-level representation from the joint observation of all neighboring agents. Second, we decompose this high-level representation into an agent-specific cognitive representation and a neighborhood-specific cognitive representation. Third, we assume that each neighborhood has a true hidden cognitive variable, then all neighboring agents learn to align their neighborhood-specific cognitive representations with this true hidden cognitive variable by variational inference. As a result, all neighboring agents will eventually form consistent neighborhood cognitions.

Due to the consistency between neighborhood-specific cognitions as well as the difference between agent-specific cognitions, the neighboring agents can achieve coordinated and still personalized policies based on the combination of both cognitions. Meanwhile, since some agents belong to multiple neighborhoods, they are able to act as a bridge for all agents. Thus, our methods can facilitate the coordination among a large number of agents at the whole team level. 

We evaluate our methods on several challenging tasks. Experiments show that they not only significantly outperform state-of-the-art MARL approaches, but also achieve greater advantages as the environment complexity grows. In addition, ablation studies and further analyses are provided for better understanding of our methods.

\section{Background}
\textbf{DEC-POMDP.} We consider a multi-agent setting that can be formulated as DEC-POMDP \cite{bernstein2002complexity}. It is formally defined as a tuple $\langle N,S,\vec{A},T,\vec{R},\vec{O},Z,\gamma \rangle$, where $N$ is the number of agents; $S$ is the set of state; $\vec{A}=A_1 \times A_2 \times ... A_N$ represents the set of \emph{joint action}, and $A_i$ is the set of \emph{local action} that agent $i$ can take; $T(s'|s,\vec{a}): S \times \vec{A} \times S \rightarrow [0,1]$ represents the state transition function; $\vec{R}=[R_1, ..., R_N]: S \times \vec{A} \rightarrow \mathbb{R}^N$ is the \emph{joint reward} function; $\vec{O} = [O_1, ..., O_N]$ is the set of \emph{joint observation} controlled by the observation function $Z: S \times \vec{A} \rightarrow \vec{O}$; $\gamma \in [0,1]$ is the \emph{discount factor}.

We focus on \textbf{\emph{fully cooperative}} settings. In a given state $s$, each agent takes an action $a_i$ based on its observation $o_i$. The joint action $\vec{a} = \langle a_{i}, \vec{a}_{-i} \rangle$ results in a new state $s'$ and a joint reward $\vec{r}=[r_1, ..., r_N]$, where $\vec{a}_{-i}$ is the joint action of teammates of agent $i$. In the following, we use $r$ to represent different $r_i$ due to $r_i=r_j$ in the fully cooperative setting. The agent aims at learning a policy $\pi_{i}(a_i|o_i): O_i \times A_i \rightarrow [0,1]$ that can maximize $\mathbb{E}[G]$ where $G$ is the \emph{discount return} defined as $G = \sum_{t=0}^{H} \gamma^{t} r^{t}$ and $H$ is the time horizon. In practice, the agent generates actions based on its observation history and some neighboring information instead of just the current observation.

\textbf{Reinforcement Learning (RL).} RL \cite{sutton1998introduction} is generally used to solve special DEC-POMDP problems where $N=1$. In practice, the Q-value (or, action-value) function is defined as $Q(s,a) = \mathbb{E}[G|S=s,A=a]$, then the optimal action can be derived by $a^* = \arg\max_a Q(s,a)$.

Q-learning \cite{tan1993multi} uses value iteration to update $Q(s,a)$. To tackle high-dimensional state and action spaces, Deep Q-network (DQN) \cite{mnih2015human} represents the Q-value function with a deep neural network $Q(s,a;w)$. DQN applies \emph{target network} and \emph{experience replay} to stabilize training and to improve data efficiency. The parameters $w$ are updated by minimising the squared TD error $\delta$:
\begin{eqnarray}
L(w) &=& \mathbb{E}_{(s,a,r,s') \sim D}[(\delta)^{2}] \label{equ:DQN1} \\
\delta &=& r + \gamma \max_{a'}Q(s',a';w^{-}) - Q(s,a;w) \label{equ:DQN2}
\end{eqnarray}
where $D$ is the replay buffer containing recent experience tuples $(s,a,r,s')$; $Q(s,a;w^{-})$ is the target network whose parameters $w^{-}$ are periodically updated by copying $w$.

Deterministic Policy Gradient (DPG) \cite{silver2014deterministic} is a special Actor-Critic algorithm where the actor adopts a deterministic policy $\mu_{\theta}: S \rightarrow A$ and the action space $A$ is continuous. Deep DPG (DDPG) \cite{duan2016benchmarking} uses deep neural networks to approximate the actor $\mu_{\theta}(s)$ and the critic $Q(s,a;w)$. Like DQN, DDPG also applies \emph{target network} and \emph{experience replay} to update the critic and actor based on the following equations:
\begin{eqnarray}
L(w) =& \hspace{-10.6em} \mathbb{E}_{(s,a,r,s') \sim D}[(\delta)^{2}] \label{equ:DPG1} \\
\delta =& \hspace{-0.5em} r + \gamma Q(s',a';w^{-})|_{a'=\mu_{\theta^{-}}(s')} - Q(s,a;w) \label{equ:DPG2} \\
\nabla_{\theta}J(\theta) =& \hspace{-1.8em} \mathbb{E}_{s \sim D}[\nabla_{\theta}\mu_\theta(s) * \nabla_{a}Q(s,a;w)|_{a=\mu_{\theta}(s)}] \label{equ:DPG3}
\end{eqnarray}

\section{Methods}
We represent a multi-agent environment as a graph $G$, where the nodes of $G$ stand for the agents, and the link between two nodes represents a relationship, e.g., a communication channel exists, between corresponding agents. The neighboring agents linked with agent $i$ are represented as $N(i)$, and each agent $j \in N(i)$ is within the \textbf{\emph{neighborhood}} of agent $i$. 

In order to study the neighborhood cognitive consistency, we define \textbf{\emph{cognition}} of an agent as its understanding of the local environment. It includes the observations of all agents in its neighborhood, as well as the high-level knowledge extracted from these observations (e.g., learned through deep neural networks). We also define \textbf{\emph{neighborhood cognitive consistency}} as that the neighboring agents have formed similar cognitions (e.g., as measured by the similar distribution of cognition variables) about their neighborhood.

Here, we exploit either Q-learning for discrete actions or DPG for continuous actions to implement the neighborhood cognitive consistency. The resulting new methods are named as NCC-Q and NCC-AC, respectively.

\subsection{Overall Design of NCC-Q}
\begin{figure*}[t]
	\centering
	\begin{subfigure}{.69\textwidth}
		\includegraphics[width=.95\textwidth]{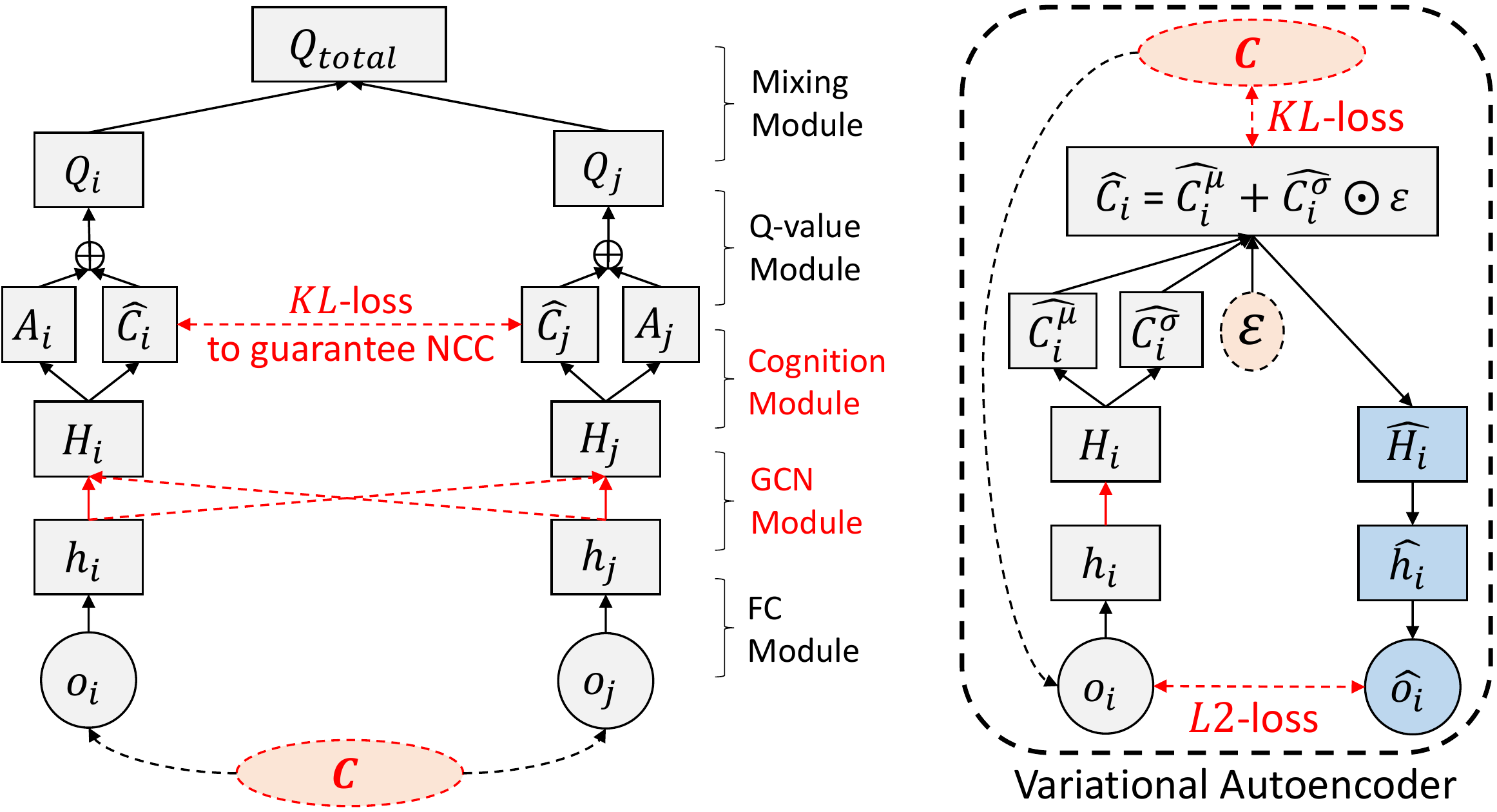}
		\caption{\textbf{Left}: the network structure of NCC-Q. \textbf{Right}: the details of a single agent $i$.}
		\label{fig:Model_NCC-Q}
	\end{subfigure}
	\begin{subfigure}{.29\textwidth}
		\includegraphics[width=.95\textwidth]{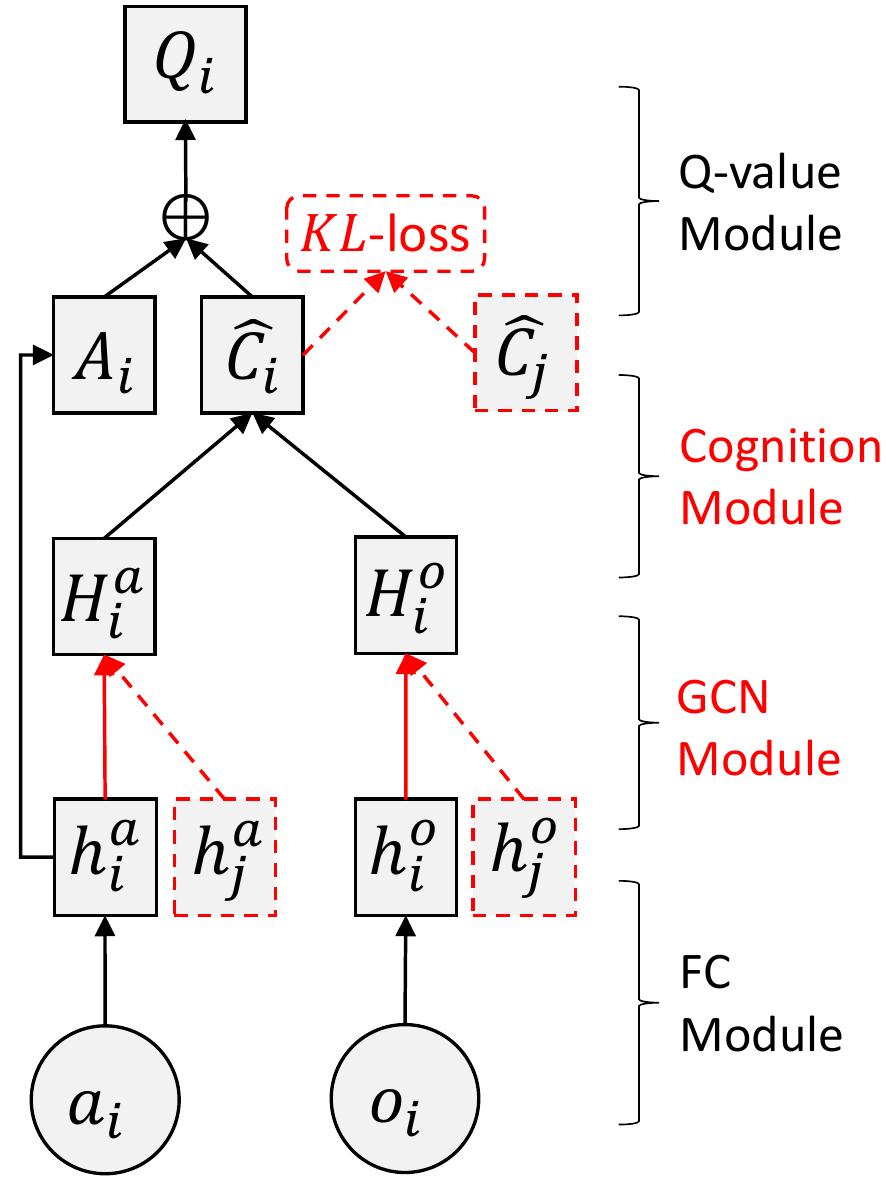}
		\caption{The critic structure of NCC-AC.}
		\label{fig:Model_NCC-AC}
	\end{subfigure}
	\caption{The network structures of NCC-Q and NCC-AC. For clarity, we show the figures assuming that agent $j$ is the only neighbor of agent $i$. The red texts indicate the main innovations to implement the neighborhood cognitive consistency. Note that $C$ is the true hidden cognitive variable that derives observations $o_i$ and $o_j$. Notations such as $h_i$, $H_i$, $A_i$, $\widehat{C_i^{\mu}}$, $\widehat{C_i^{\sigma}}$ and $\widehat{C_i}$ represent the hidden layers of the deep neural networks. $\varepsilon$ is a random sample from a unit Gaussian, i.e., $\varepsilon \sim N(0,1)$.}
	\label{fig:Model_NCC-MARL}
\end{figure*}

The network structure of NCC-Q consists of five modules as shown on the left of Figure \ref{fig:Model_NCC-Q}.

(1) The fully-connected (FC) module encodes the local observation $o_i$ into a low-level cognition vector $h_i$, which only contains agent-specific information.

(2) The graph convolutional network (GCN) module aggregates all $h_i$ within the neighborhood, and further extracts a high-level cognition vector $H_i$ by:
\begin{eqnarray}
H_i=\sigma (W \Sigma_{j \in N(i) \cap \{i\}}  \frac{h_j}{\sqrt{|N(j)||N(i)|}})
\end{eqnarray}
As the equation shows, all neighboring agents (including $i$ and $N(i)$) use the same parameters $W$ to generate $H_i$. This parameter sharing idea reduces overfitting rick and makes the method more robust. We also normalize features $h_j$ by $\sqrt{|N(j)||N(i)|}$ instead of by a fixed $|N(i)|$, since it is a more adaptive way to down-weight high-degree neighbors.

(3) The cognition module decomposes $H_i$ into two branches: $A_i$ and $\widehat{C_i}$. In our design, $A_i$ represents an agent-specific cognition, and it should capture the local knowledge about the agent's own situation. In contrast, $\widehat{C_i}$ represents the neighborhood-specific cognition, and it should extract general knowledge about the neighborhood. By decomposing $H_i$ into different knowledge branches, the neighborhood cognitive consistency constraints can be explicitly imposed on $\widehat{C_i}$ to achieve more efficient cooperation among agents. The implementation details will be covered in next section.

(4) The Q-value module adopts element-wise summation to aggregate $A_i$ and $\widehat{C_i}$. It then generates the Q-value function $Q_i(o_i,a_i)$ based on the result of summation.

(5) The mixing module sums up all $Q_i(o_i,a_i)$ to generate a joint Q-value function $Q_{total}(\vec{o},\vec{a})$, which is shared by all agents. This layer resembles the state-of-the-art VDN \cite{sunehag2018value} and QMIX \cite{rashid2018qmix}. It not only provides many benefits like team utility decomposition, but also guarantees a fair comparison.

\subsection{Neighborhood Cognitive Consistency}
In the GCN module, all neighboring agents adopt the same parameters $W$ to extract $H_i$. In this way, they are \emph{more likely} to form consistent neighborhood-specific cognitions $\widehat{C_i}$. However, this cannot be \emph{guaranteed} only by a single GCN module. To this end, we propose a more principle and practical implementation of neighborhood cognitive consistency. The idea is based on two innovative assumptions.

\textbf{Assumption 1.} For each neighborhood, there is a \textbf{\emph{true hidden cognitive variable $\bm{C}$}} to derive the observation $o_j$ of each agent $j \in N(i) \cap \{i\}$.

\begin{figure}[t]
	\centering
	\includegraphics[width=.95\columnwidth]{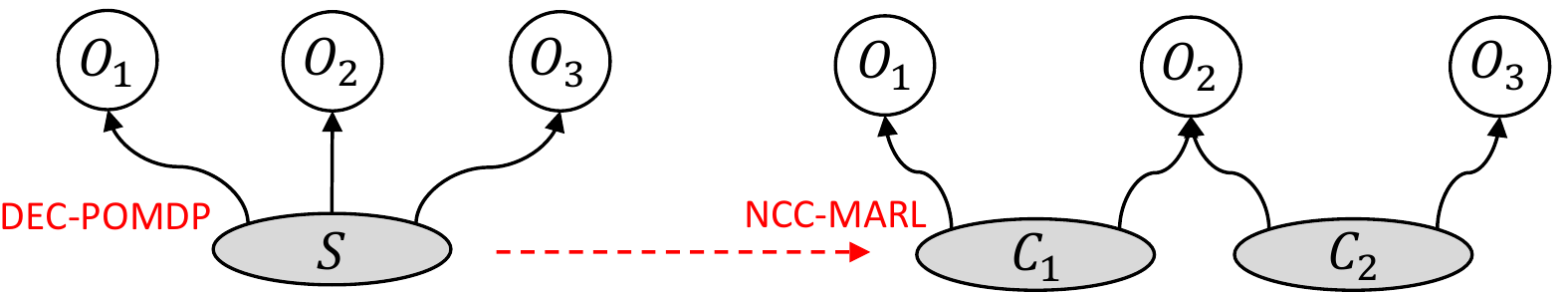}
	\caption{In NCC-MARL, the observations $O_i$ are generated based on the hidden cognitive variable $C_i$ instead of global state $S$. Here, agent $2$ belongs to two neighborhoods.}
	\label{fig:nccmarl}
\end{figure}

To make this assumption easier to understand, we show the paradigms of DEC-POMDP and our NCC-MARL in Figure \ref{fig:nccmarl}. NCC-MARL decomposes the global state $S$ into several hidden cognitive variables $C_k$, and the observations $O_i$ are derived based on (possibly several) $C_k$. This assumption is very flexible. In small-scale settings where there is only one neighborhood, $C$ is reduced to be equivalent with $S$, then DEC-POMDP can be viewed as a special case of NCC-MARL. In large-scale settings, decomposing $S$ into individual cognitive variables for each neighborhood is more in line with the reality: the neighboring agents usually have closer relationship and similar perceptions, so they are more likely to form consistent cognition about their local environments. In contrast, describing all agents by a single state is inaccurate and unrealistic, because agents who are far away from each other usually have very different observations.

\textbf{Assumption 2.} If the neighboring agents can recover the true hidden cognitive variable $C$, they will eventually form consistent neighborhood cognitions and thus achieve better cooperations. In other words, the \emph{learned} cognitive variable $\widehat{C_i}$ should be similar to the \emph{true} cognitive variable $C$.

We formally formulate the above assumption as follows. Supposing each agent $i$ can only observe $o_i$ \footnote{Note that agent $i$ can also observe $h_j$ of its neighboring agents due to the GCN module. We omit $h_j$ to make the notations concise.}, there exists a hidden process $p(o_i|C)$, and we would like to infer $C$ by:
\begin{eqnarray}
p(C|o_i) = \frac{p(o_i|C)p(C)}{p(o_i)} = \frac{p(o_i|C)p(C)}{\int p(x|C)p(C)dC}
\end{eqnarray}

However, directly computing the above equation is quite difficult, so we approximate $p(C|o_i)$ by another distribution $q(C|o_i)$ that has a tractable distribution. The only restriction is that $q(C|o_i)$ needs to be similar to $p(C|o_i)$. We achieve this by minimizing the following KL divergence:
\begin{eqnarray}
\min KL(q(C|o_i)||p(C|o_i))
\end{eqnarray}
which equals to maximize the following \cite{blei2017variational}:
\begin{eqnarray}
\max \mathbb{E}_{q(C|o_i)}\log p(o_i|C) - KL(q(C|o_i)||p(C)) \label{equ:VAE-true}
\end{eqnarray}

In the above equation, the first term represents the reconstruction likelihood, and the second term ensures that our learned distribution $q(C|o_i)$ is similar to the true prior distribution $p(C)$. This can be modelled by a variational autoencoder (VAE) as shown on the right of Figure \ref{fig:Model_NCC-Q}. The encoder of this VAE learns a mapping $q(\widehat{C_i}|o_i;w)$ from $o_i$ to $\widehat{C_i}$. Specifically, rather than directly outputting $\widehat{C_i}$, we adopt the ``reparameterization trick'' to sample $\varepsilon$ from a unit Gaussian, and then shift the sampled $\varepsilon$ by the latent distribution's mean $\widehat{C_i^{\mu}}$ and scale it by the latent distribution's variance $\widehat{C_i^{\sigma}}$. That is to say, $\widehat{C_i} = \widehat{C_i^{\mu}} + \widehat{C_i^{\sigma}} \odot \varepsilon$ where $\varepsilon \sim N(0,1)$. The decoder of this VAE learns a mapping $p(\widehat{o_i}|\widehat{C_i};w)$ from $\widehat{C_i}$ back to $\widehat{o_i}$. The loss function to train this VAE is:
\begin{eqnarray}
\min L2(o_i,\widehat{o_i};w) + KL(q(\widehat{C_i}|o_i;w)||p(C)) \label{equ:VAE-appr}
\end{eqnarray} 

Comparing Equation \ref{equ:VAE-appr} with Equation \ref{equ:VAE-true}, it can be seen that the learned $q(\widehat{C_i}|o_i;w)$ is used to approximate the true $q(C|o_i)$. In our method, all neighboring agents are trained using the same $p(C)$, therefore they will eventually learn a cognitive variable $\widehat{C_i}$ that is aligned well with the true hidden cognitive variable $C$, namely, they finally achieve consistent cognitions at the neighborhood level.

\subsection{Training Method of NCC-Q}
NCC-Q is trained by minimizing two loss functions. First, a temporal-difference loss (TD-loss) is shared by all agents:
\begin{eqnarray}
L^{td}(w) &=& \mathbb{E}_{(\vec{o},\vec{a},r,\vec{o}')} [(y_{total}-Q_{total}(\vec{o},\vec{a};w))^2] \label{equ:TD-loss} \\
y_{total} &=& r + \gamma \max_{\vec{a}'}Q_{total}(\vec{o}',\vec{a}';w^{-})
\end{eqnarray}
This is analogous to the standard DQN loss shown in Equation \ref{equ:DQN1} and \ref{equ:DQN2}. It encourages all agents to cooperatively produce a large $Q_{total}$, and thus ensures good agent cooperation at the whole team level as the training goes on.

Second, a cognitive-dissonance loss (CD-loss) is specified for each agent $i$:
\begin{eqnarray}
L^{cd}_{i}(w) \hspace{-0.2em} = \hspace{-0.2em} \mathbb{E}_{o_i} [L2(o_i,\widehat{o_i};w) + KL(q(\widehat{C_i}|o_i;w)||p(C))] \label{equ:cd-loss}
\end{eqnarray}
This is a mini-batch version of Equation \ref{equ:VAE-appr}. It ensures that cognitive consistency and good agent cooperation can be achieved at the neighborhood level as the training goes on.

The total loss is a combination of Equation \ref{equ:TD-loss} and \ref{equ:cd-loss}:
\begin{eqnarray}
L^{total}(w) = L^{td}(w) + \alpha \Sigma_{i=1}^{N} L^{cd}_{i}(w) \label{total-loss}
\end{eqnarray}

Nevertheless, there are two remaining questions about the CD-loss $L^{cd}_{i}(w)$. (1) The true \textbf{\emph{hidden}} cognitive variable $C$ and its distribution $p(C)$ are unknown. (2) If there are multiple agent neighborhoods, how to choose a suitable $p(C)$ for each neighborhood.

In cases that there is only one neighborhood (e.g., the number of agents is small), we assume that $p(C)$ follows a unit Gaussian distribution, which is commonly used in many variational inference problems. However, if there are more neighborhoods, it is neither elegant nor appropriate to apply the same $p(C)$ for all neighborhoods. In practice, we find that the neighboring agents' cognitive distribution $q(\widehat{C_j}|o_j;w)$ is a good surrogate for $p(C)$. Specifically, we approximate the cognitive-dissonance loss by:
\begin{eqnarray}
L^{cd}_{i}(w) =& \hspace{-1.0em} \mathbb{E}_{o_i} [L2(o_i,\widehat{o_i};w) + KL(q(\widehat{C_i}|o_i;w)||p(C))]  \nonumber \\
\approx& \hspace{-10.5em} \mathbb{E}_{o_i} [L2(o_i,\widehat{o_i};w) + \label{equ:CD-loss} \\
& \hspace{-1.0em} \frac{1}{|N(i)|} \Sigma_{j \in N(i)} KL(q(\widehat{C_i}|o_i;w)||q(\widehat{C_j}|o_j;w))] \nonumber
\end{eqnarray}

This approximation punishes any neighboring agents $i$ and $j$ with large cognitive dissonance, namely, with a large $KL(q(\widehat{C_i}|o_i;w)||q(\widehat{C_j}|o_j;w))$. Thus, it also guarantees the neighborhood cognitive consistency. The key intuition behind this approximation is that we do not care about whether the neighboring agents converge to a specific cognitive distribution $p(C)$, as long as they have formed consistent cognitions as measured by a small $KL(q(\widehat{C_i}|o_i;w)||q(\widehat{C_j}|o_j;w))$.

\subsection{NCC-AC}
NCC-AC adopts an actor-critic architecture like the state-of-the-art MADDPG \cite{Lowe2017Multi} and ATT-MADDPG \cite{mao2019modelling} to guarantee fair comparisons. Specifically, each agent $i$ adopts an independent actor $\mu_{\theta_i}(o_i)$, which is exactly the same as \cite{Lowe2017Multi,mao2019modelling}. For the critic $Q_i(\langle o_i, a_{i} \rangle,\vec{o}_{-i},\vec{a}_{-i};w_{i})$ shown in Figure \ref{fig:Model_NCC-AC}, two major innovations, namely the GCN module and cognition module, are designed for achieving neighborhood cognitive consistency and thus good agent cooperation. Since the two modules share the same idea as our NCC-Q introduced before, we will not repeat them.

There are some details. (1) Instead of taking action and observation as a whole, the GCN module integrates the action and observation \emph{separately} as shown by the two branches (i.e., $H^a_i$ and $H^o_i$) in Figure \ref{fig:Model_NCC-AC}. (2) We directly apply $h^a_i$ to generate the agent-specific cognition variable $A_i$ as shown by the leftmost shortcut connection in Figure \ref{fig:Model_NCC-AC}. These designs are helpful for performance empirically.

Like NCC-Q, the critic of NCC-AC is trained by minimizing the combination of $L^{td}_{i}(w_i)$ and $L^{cd}_{i}(w_i)$ as follows:
\begin{eqnarray}
L^{total}_{i}(w_i) =& \hspace{-9.8em}  L^{td}_{i}(w_i) + \alpha L^{cd}_{i}(w_i) \\
L^{td}_{i}(w_i) =& \hspace{-5.2em} \mathbb{E}_{(o_i,\vec{o}_{-i},a_i,\vec{a}_{-i},r,o'_i,\vec{o}'_{-i}) \sim D}[(\delta_i)^{2}] \\
\delta_i =& \hspace{-0.8em} r + \gamma Q_i(\langle o'_i,a'_i \rangle, \vec{o}'_{-i},\vec{a}'_{-i};w_i^{-})|_{a'_j=\mu_{\theta_j^{-}}(o'_j)} \nonumber \\
& \hspace{-5.8em} - Q_{i}(\langle o_i,a_{i} \rangle,\vec{o}_{-i},\vec{a}_{-i};w_{i}) \\
L^{cd}_{i}(w_i) \approx& \hspace{-2.9em} \mathbb{E}_{o_i} [L2(o_i,\widehat{o_i};w_i) + L2(a_i,\widehat{a_i};w_i) + \\
& \hspace{-5.0em} \frac{1}{|N(i)|} \Sigma_{j \in N(i)} KL(q(\widehat{C_i}|o_i,a_i;w_i)||q(\widehat{C_j}|o_j,a_j;w_j))] \nonumber
\end{eqnarray}

As for the actor of NCC-AC, we extend Equation \ref{equ:DPG3} into multi-agent formulation as follows:
\begin{eqnarray}
\nabla_{\theta_i}J(\theta_i) =& \hspace{-6.8em} \mathbb{E}_{(o_i,\vec{o}_{-i}) \sim D}[\nabla_{\theta_i}\mu_{\theta_i}(o_i) * \nonumber \\
& \hspace{-0.8em} \nabla_{a_i}Q_{i}(\langle o_i,a_{i} \rangle,\vec{o}_{-i},\vec{a}_{-i};w_{i})|_{a_j=\mu_{\theta_j}(o_j)}] \label{equ:ATT-MADDPG3}
\end{eqnarray}

\section{Experimental Evaluations}
\subsection{Environments}
For continuous actions, we adopt the packet routing tasks proposed by ATT-MADDPG \cite{mao2019modelling} to ensure fair comparison. For discrete actions, we test wifi configuration and Google football environments. In these environments, the natural topology between agents can be used to form neighborhoods, therefore we can evaluate our methods without disturbing by neighborhood formation, which is not the focus of this work.

\begin{figure*}[t]
	\centering
	\begin{subfigure}{.24\textwidth}
		\centering
		\includegraphics[width=.8\textwidth]{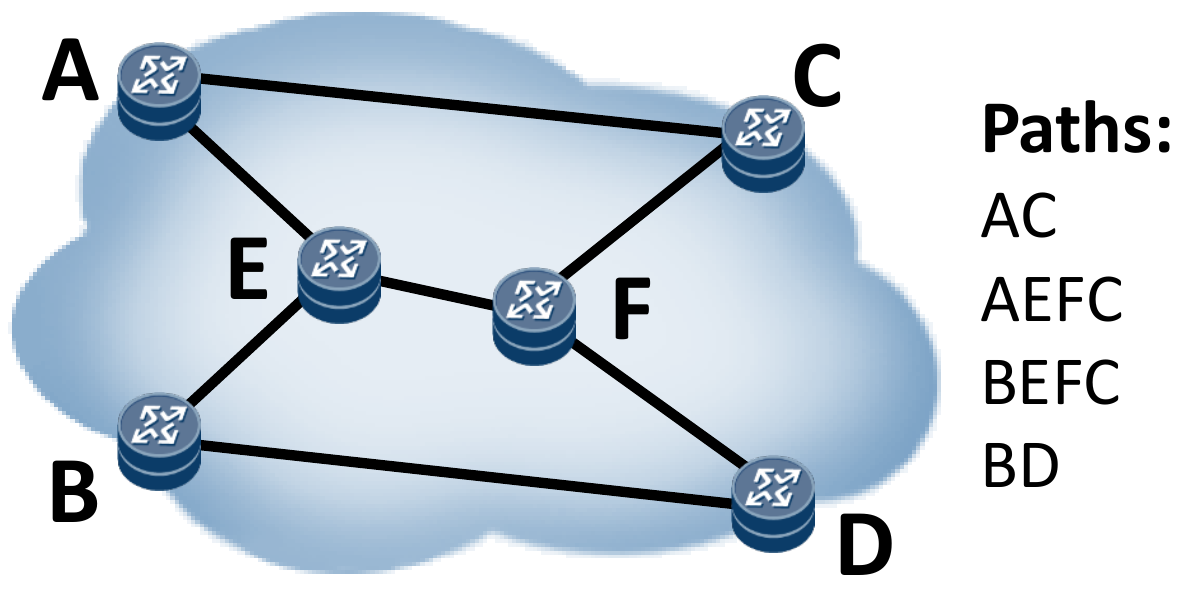}
		\caption{6 routers and 4 paths.}
		\label{fig:routing-small}
	\end{subfigure}
	\begin{subfigure}{.24\textwidth}
		\centering
		\includegraphics[width=.8\textwidth]{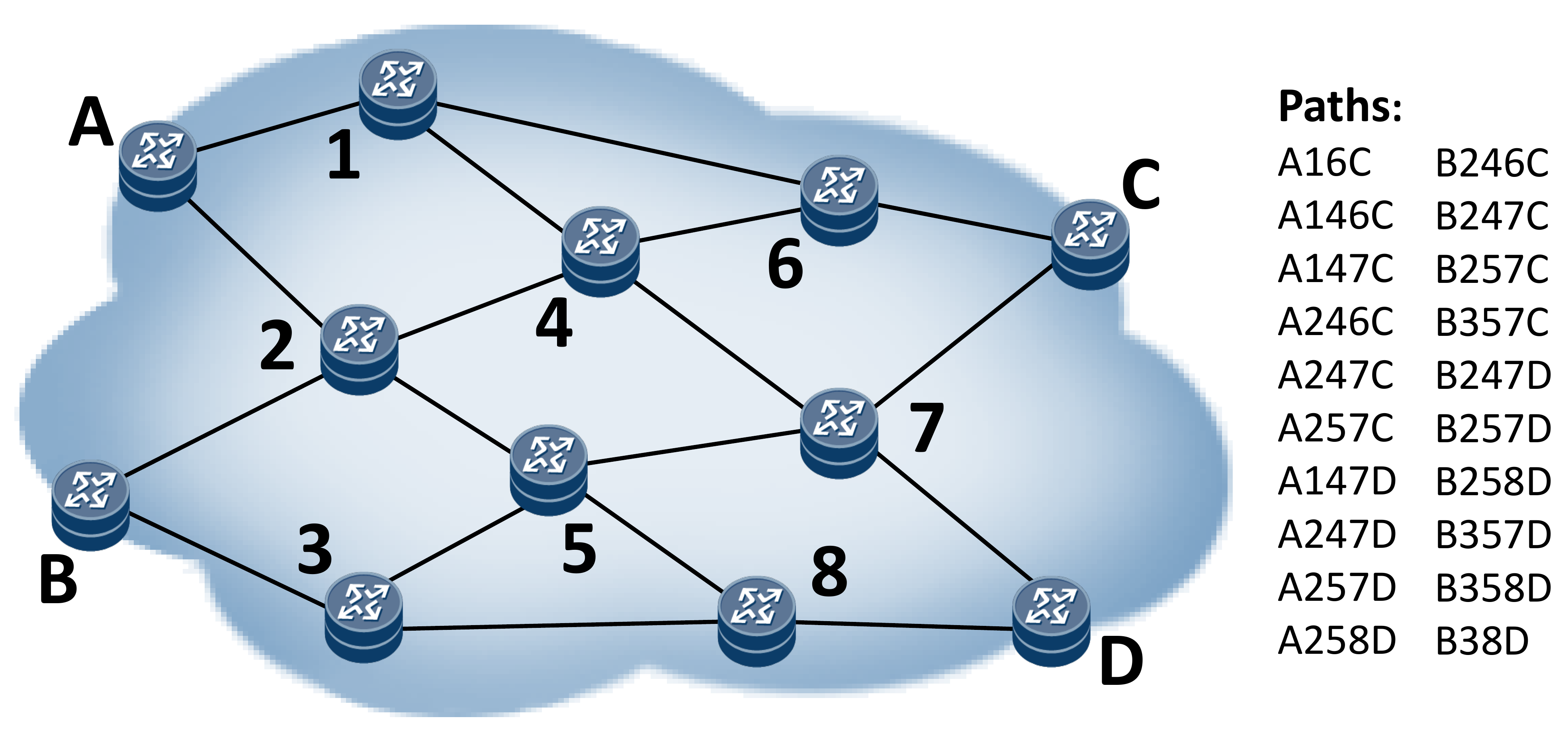}
		\caption{12 routers and 20 paths.}
		\label{fig:routing-middle}
	\end{subfigure}
	\begin{subfigure}{.24\textwidth}
		\centering
		\includegraphics[width=.7\textwidth]{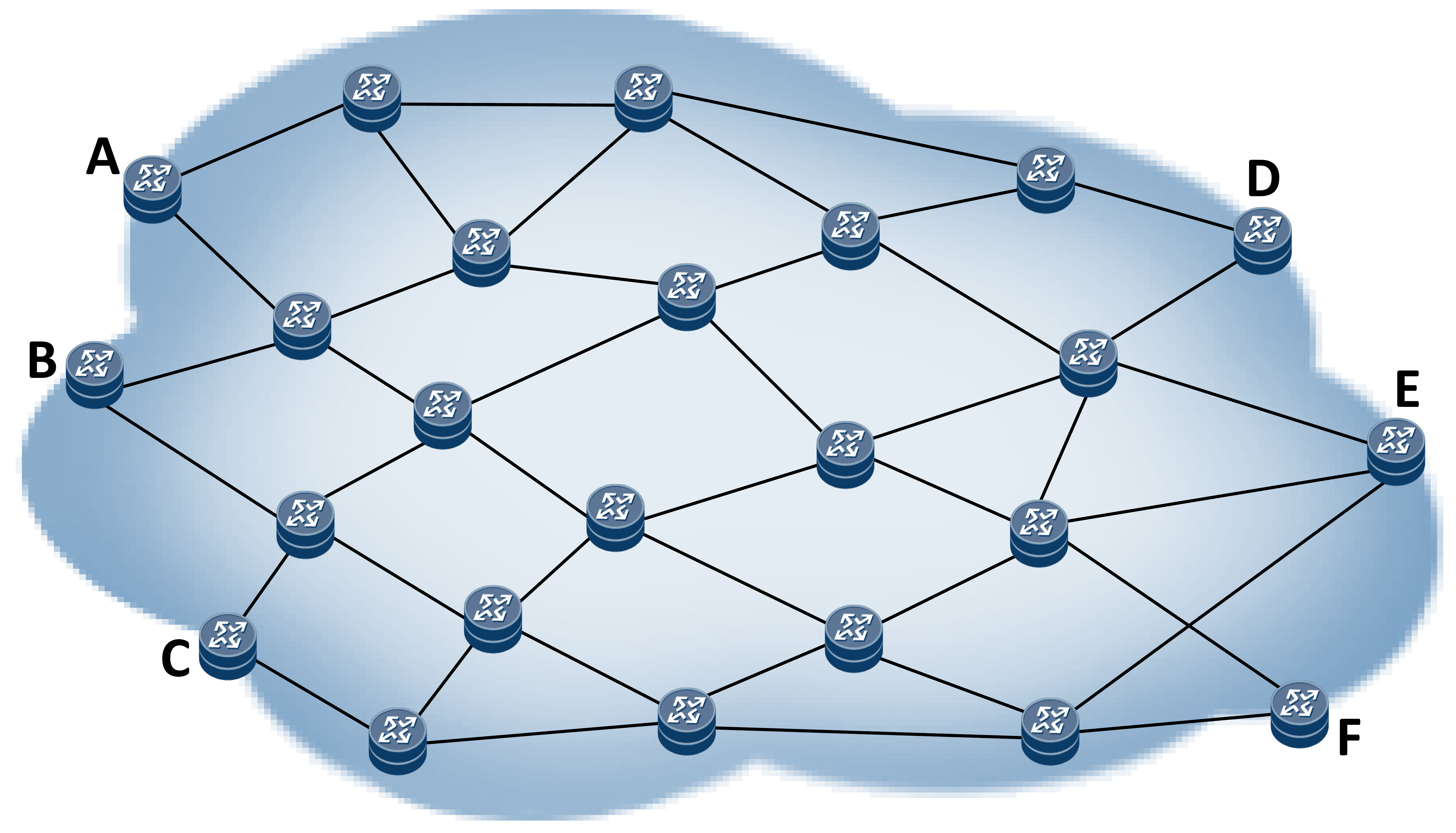}
		\caption{24 routers and 128 paths!}
		\label{fig:routing-large}
	\end{subfigure}
	\begin{subfigure}{.24\textwidth}
		\centering
		\includegraphics[width=.8\textwidth]{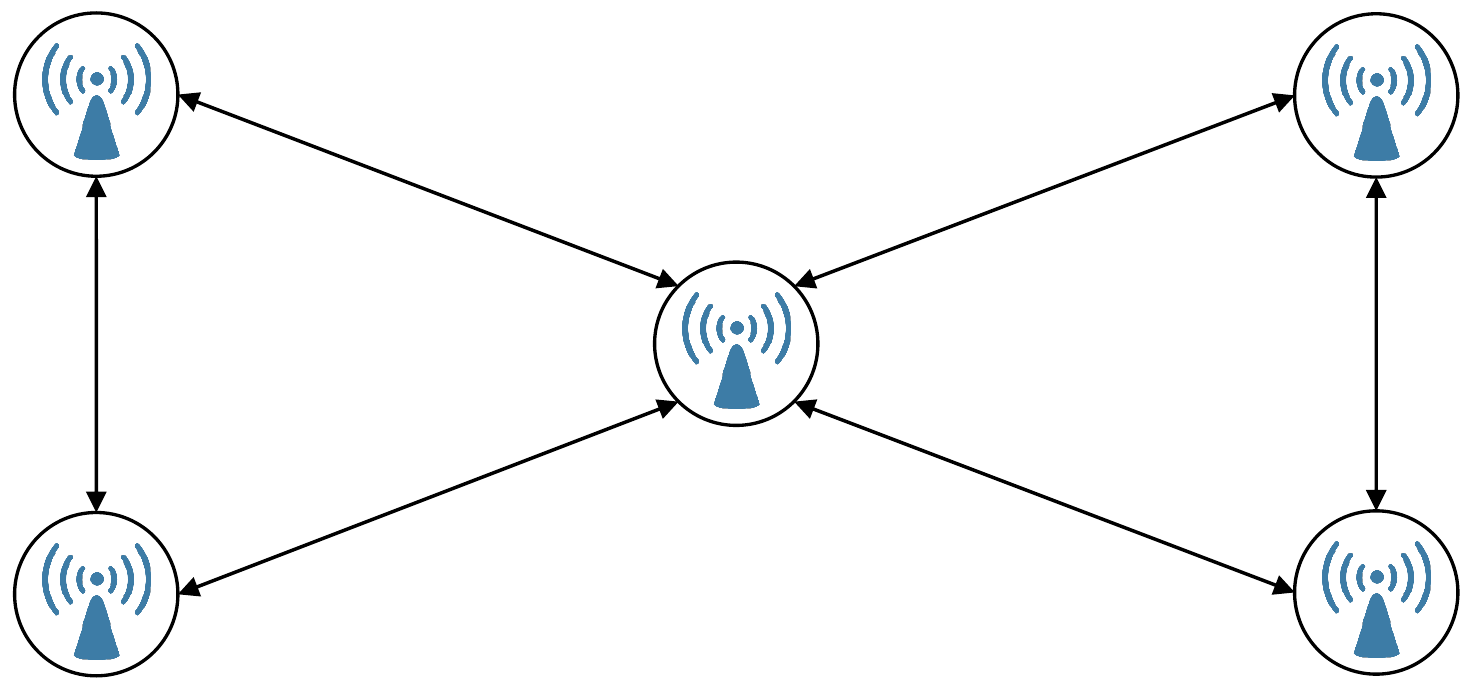}
		\caption{5 APs with 12 channels.}
		\label{fig:wifi5}
	\end{subfigure}
	\begin{subfigure}{.24\textwidth}
		\centering
		\includegraphics[width=.7\textwidth]{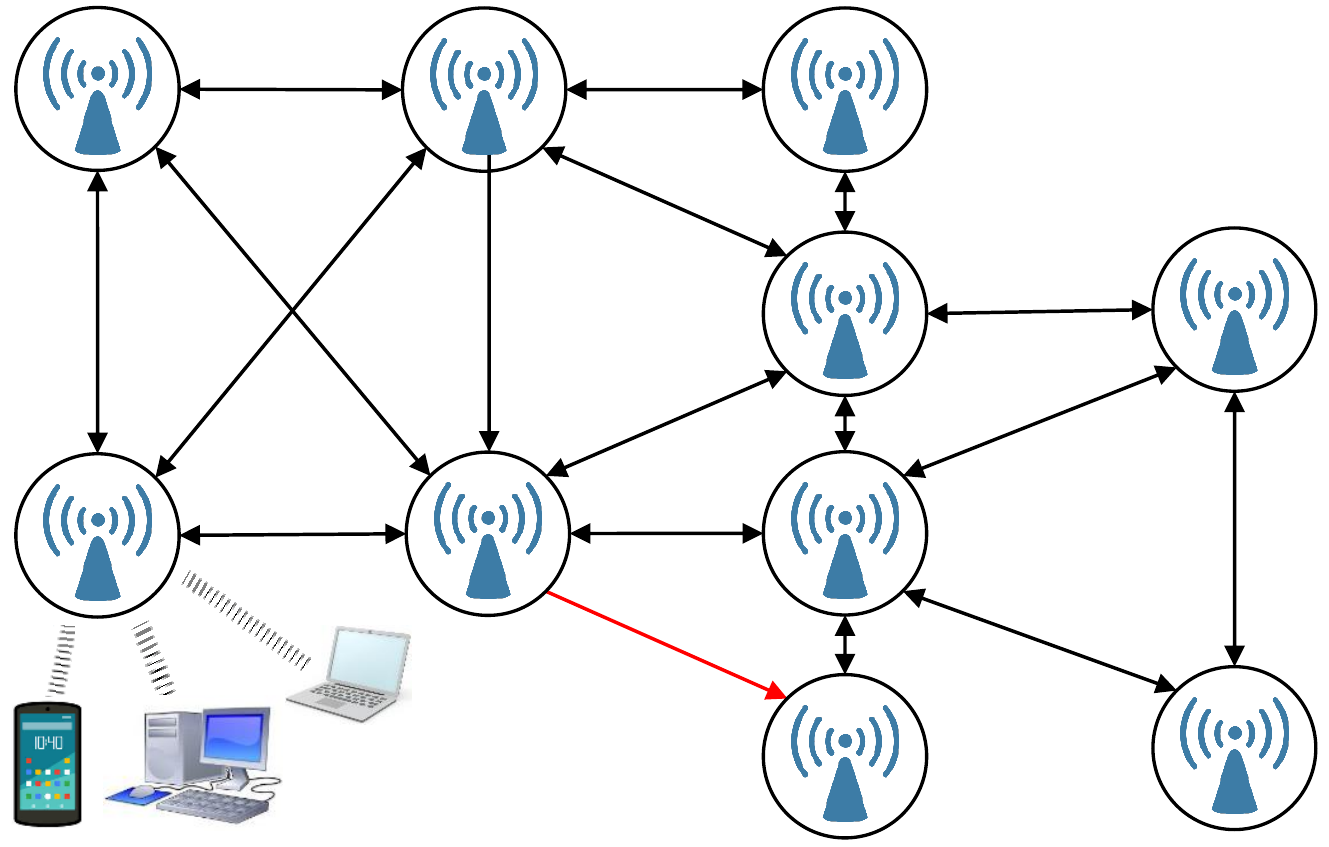}
		\caption{10 APs with 35 channels.}
		\label{fig:wifi10}
	\end{subfigure}
	\begin{subfigure}{.24\textwidth}
		\centering
		\includegraphics[width=.6\textwidth]{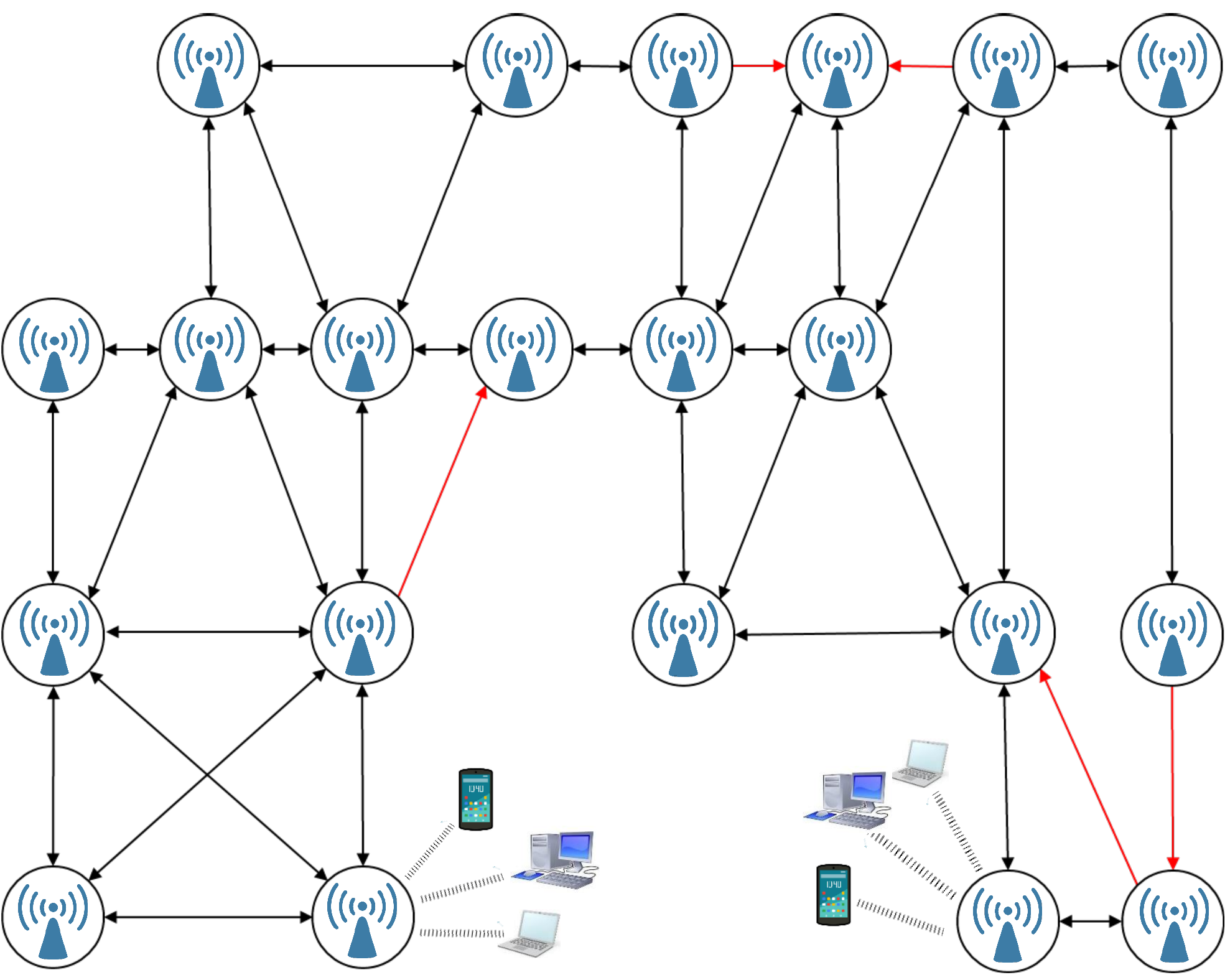}
		\caption{21 APs with 70 channels!}
		\label{fig:wifi21}
	\end{subfigure}
	\begin{subfigure}{.24\textwidth}
		\centering
		\includegraphics[width=.7\textwidth]{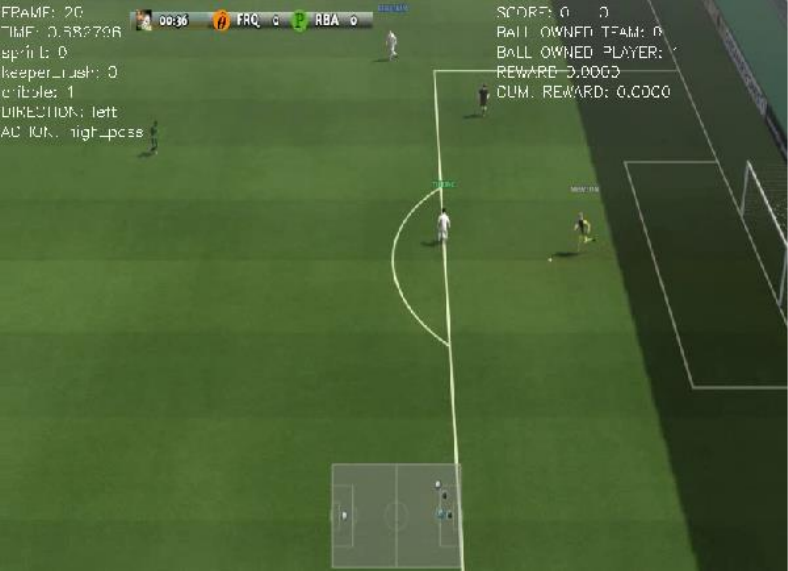}
		\caption{2-vs-2 scenario.}
		\label{fig:football2}
	\end{subfigure}
	\begin{subfigure}{.24\textwidth}
		\centering
		\includegraphics[width=.7\textwidth]{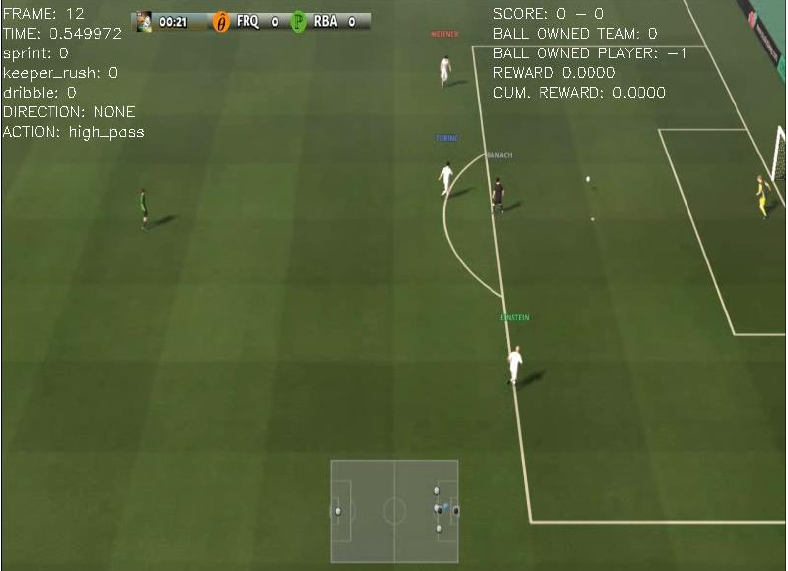}
		\caption{3-vs-2 scenario.}
		\label{fig:football3}
	\end{subfigure}
	\caption{The evaluation environments that are developed based on real-world scenarios. (a-c): The small, middle and large packet routing topologies. (d-f): The small, middle and large wifi configuration topologies. (g-h): The Google football tasks.}
	\label{fig:Environments}
\end{figure*}

\textbf{Packet Routing.} Due to space limitation, we refer the readers to \cite{mao2019modelling} for the background. Here, we only introduce the problem definition. As shown in Figure 3(a-c), the routers are controlled by our NCC-AC model, and they try to learn a good flow splitting policy to minimize the \emph{Maximum Link Utilization in the whole network (MLU)}. The intuition behind this objective is that high link utilization is undesirable for dealing with bursty traffic. For each router, the \textbf{\emph{observation}} includes the flow demands in its buffers, the estimated link utilization of its direct links during the last ten steps, the average link utilization of its direct links during the last control cycle, and the latest action taken by the router. The \textbf{\emph{action}} is the flow rate assigned to each available path. The \textbf{\emph{reward}} is $1-MLU$ as we want to minimize \emph{MLU}.

\textbf{Wifi Configuration.} The cornerstone of any wireless network is the access point (AP). The primary job of an AP is to broadcast a wireless signal that computers can detect and tune into. The power selection for AP is tedious, and the AP behaviors differ greatly in various scenarios. The current optimization is highly depending on human expertise, but the expert knowledge cannot deal with interference among APs and fails to handle dynamic environments.

In the tasks shown in Figure 3(d-f), the APs are controlled by our NCC-Q model. They aim at learning a good power configuration policy to maximize the Received Signal Strength Indication (RSSI). In general, larger RSSI indicates better signal quality. For each AP, the \textbf{\emph{observation}} includes radio frequency, bandwidth, the rate of package loss, the number of band, the current number of users in one specific band, the number of download bytes in ten seconds, the upload coordinate speed (Mbps), the download coordinate speed and the latency. An \textbf{\emph{action}} is the actual power value specified by an integer between ten and thirty. The \textbf{\emph{reward}} is RSSI and the goal is to maximize accumulated RSSI.

\textbf{Google Football.} In this environment, agents are trained to play football in an advanced, physics-based 3D simulator. The environment is very challenging due to high complexity and inner random noise. For each agent, the \textbf{\emph{observation}} is the game state encoded with 115 floats, which includes coordinates of players/ball, player/ball directions and so on. There are \textbf{\emph{21 discrete actions}} including move actions (up, down, left, right), different ways to kick the ball (short, long and high passes, shooting), sprint and so on. The \textbf{\emph{reward}} is $+1$ when scoring a goal, and $-1$ when conceding one to the opposing team. We refer the readers to \cite{kurach2019google} for the details.

We evaluate NCC-Q with two standard scenarios shown in Figure 3(g-h). In the 2-vs-2 scenario, \emph{two} of our players try to score from the edge of the box; one is on the side with the ball, and next to a defender; the other is at the center, facing the opponent keeper. In the 3-vs-2 scenario, \emph{three} of our players try to score from the edge of the box; one on each side, and the other at the center; initially, the player at the center has the ball, and is facing the defender; there is also an opponent keeper.

\begin{figure*}[t]
	\centering
	\begin{subfigure}{.33\textwidth}
		\includegraphics[width=.95\textwidth]{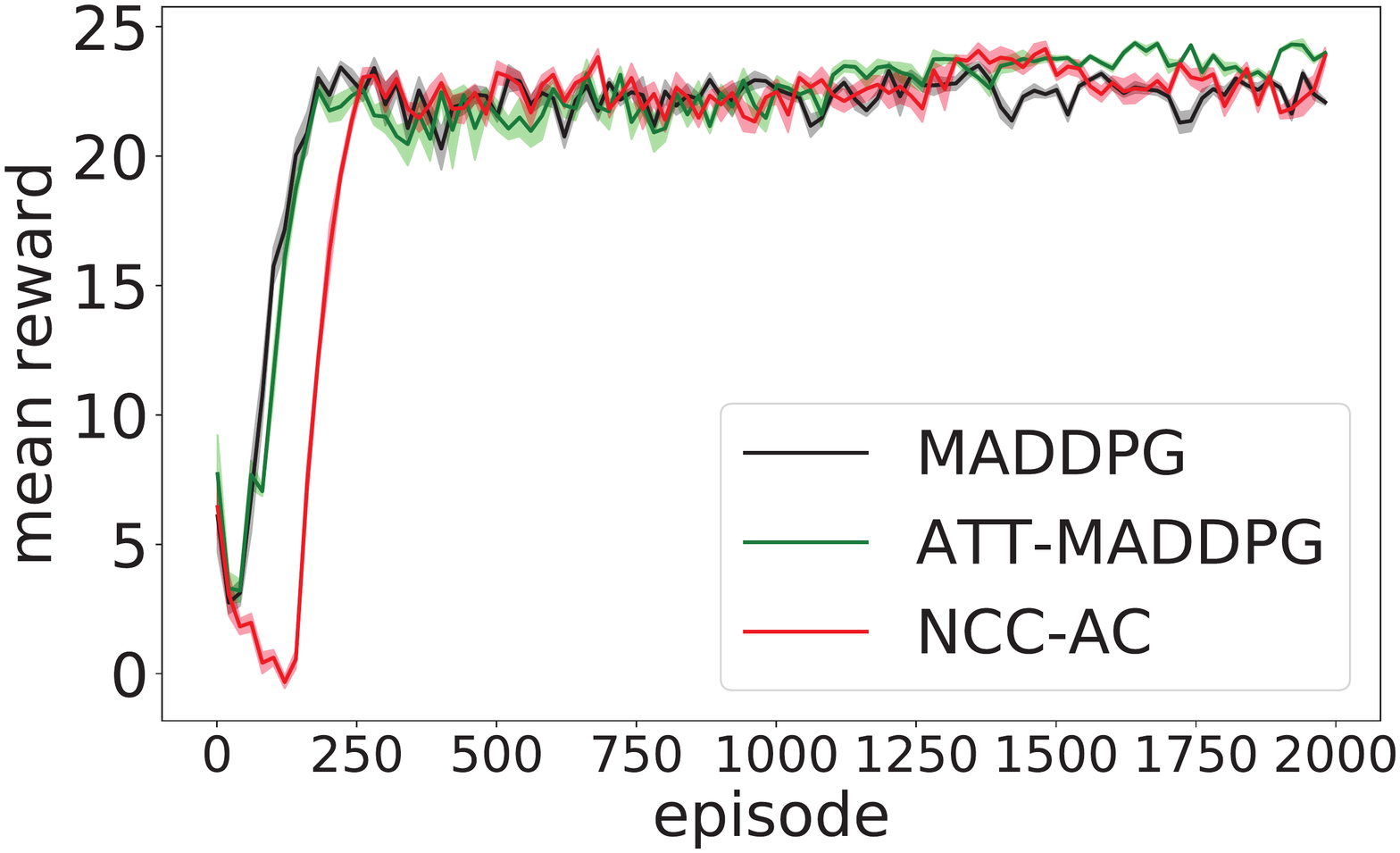}
		\caption{Small topology: 6 routers, 4 paths.}
		\label{fig:small}
	\end{subfigure}
	\begin{subfigure}{.33\textwidth}
		\includegraphics[width=.95\textwidth]{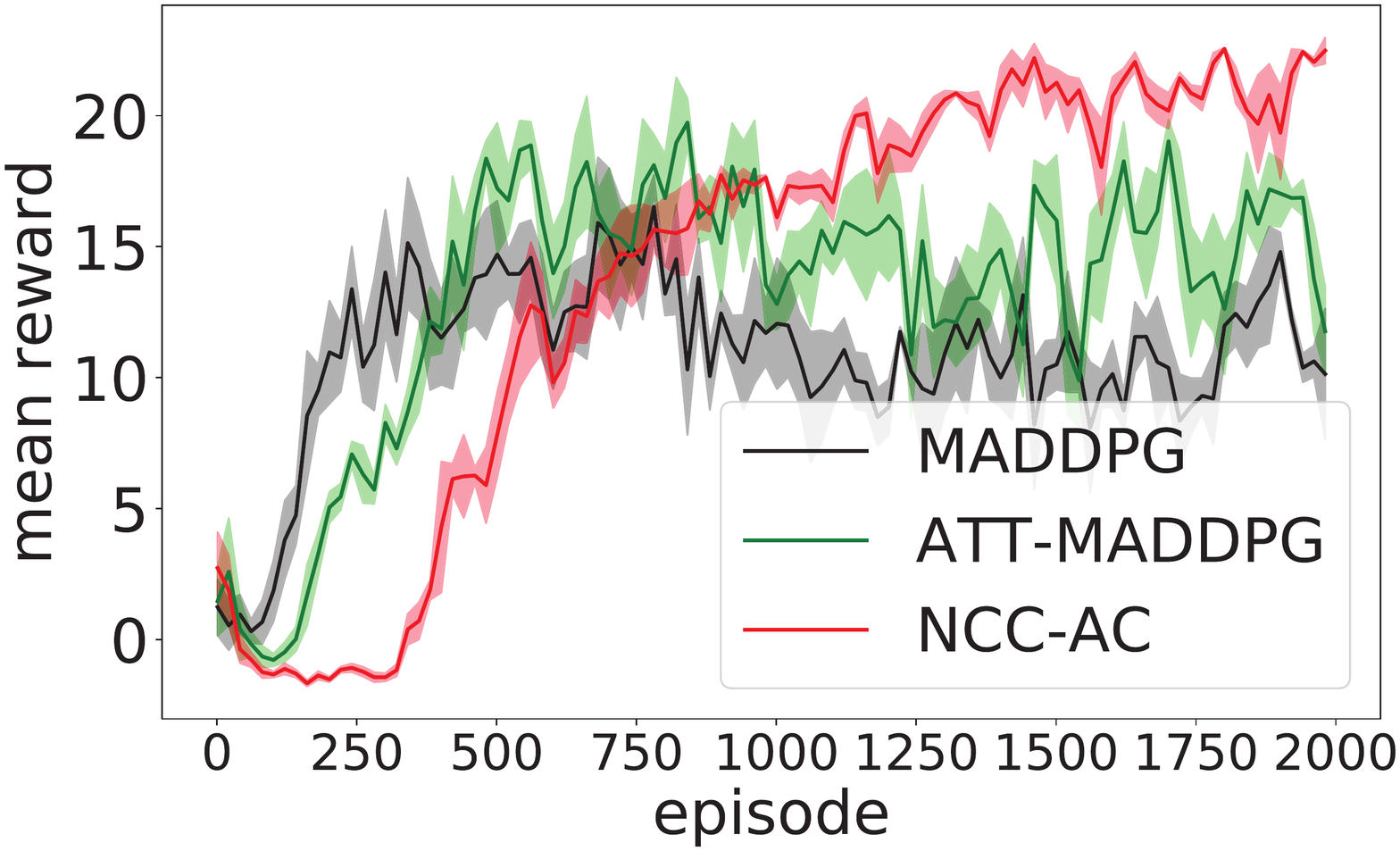}
		\caption{Middle topology: 12 routers, 20 paths.}
		\label{fig:large}
	\end{subfigure}
	\begin{subfigure}{.33\textwidth}
		\includegraphics[width=.95\textwidth]{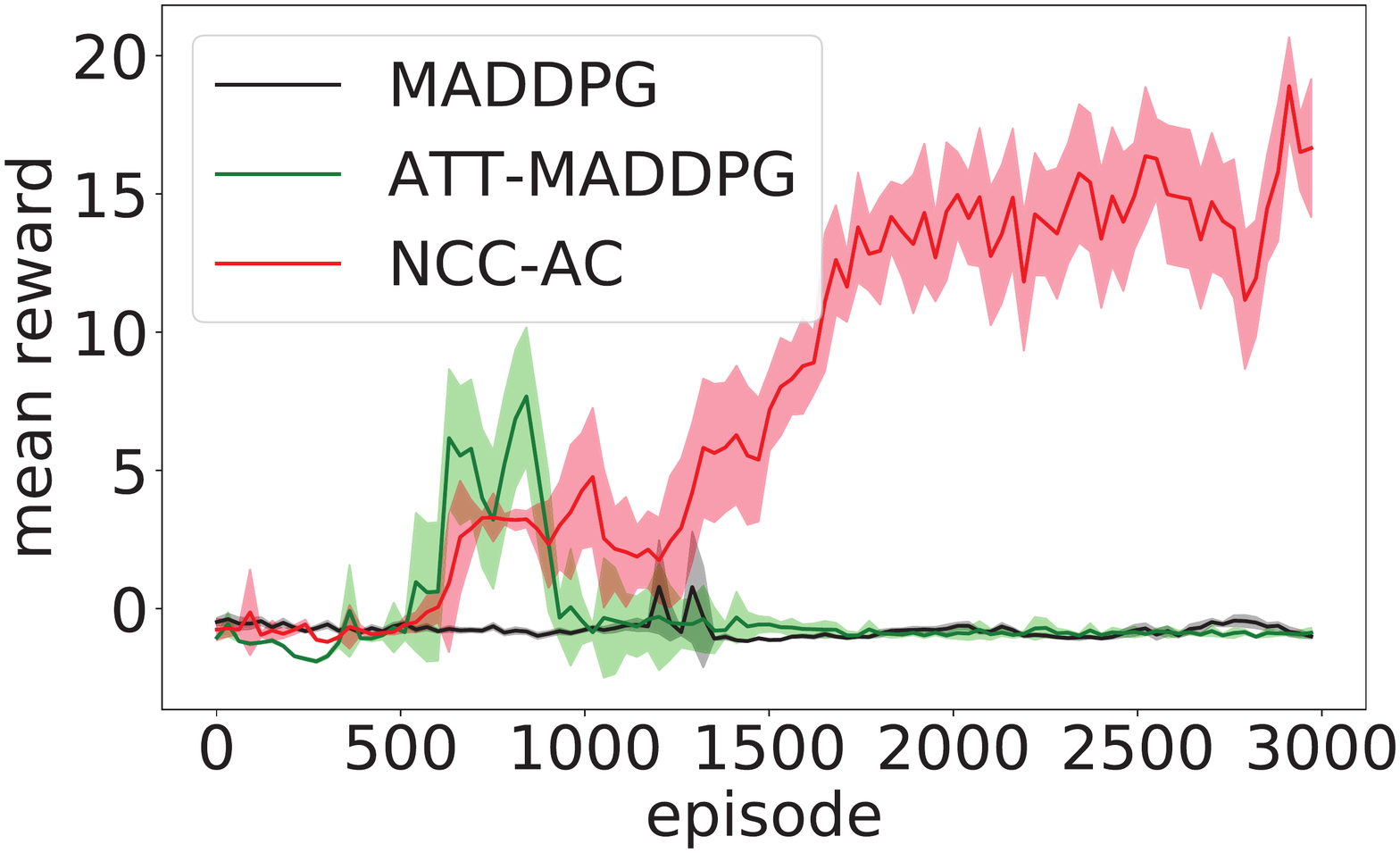}
		\caption{Large topology: 24 routers, 128 paths!}
		\label{fig:huge}
	\end{subfigure}
	\caption{The average results of different packet routing scenarios.}
	\label{fig:PacketRoutingResults}
\end{figure*}

\subsection{Baselines}
For continuous actions, we compare NCC-AC with the state-of-the-art MADDPG \cite{Lowe2017Multi} and ATT-MADDPG \cite{mao2019modelling}. All the three methods apply an independent actor for each agent. The differences are that MADDPG adopts a fully-connected network as centralized critic; ATT-MADDPG applies an attention mechanism to enhance centralized critic; NCC-AC utilizes GCN and VAE to design a centralized critic with neighborhood cognitive consistency.

For discrete actions, we compare NCC-Q with the state-of-the-art VDN \cite{sunehag2018value} and QMIX \cite{rashid2018qmix}. All the three methods generate a single Q-value function $Q_i$ for each agent $i$, and then mix all $Q_i$ into a $Q_{total}$ shared by all agents. The differences are that NCC-Q and VDN merge all $Q_i$ linearly, but QMIX nonlinearly by ``Mixing Network''; at the same time, NCC-Q applies GCN and VAE to ensure neighborhood cognitive consistency. In addition, the Independent DQN (IDQN) \cite{tampuu2017multiagent} and DGN \cite{jiang2018graph} are compared. IDQN learns an independent Q-value function $Q_i$ for each agent $i$ without mixing them. DGN adopts GCN with multi-head attention and parameter sharing method (i.e., the same set of network parameters are shared by all agents) to achieve coordination at the whole team level.

For the ablation study, we compare NCC-AC with Graph-AC and GCC-AC, and compare NCC-Q with Graph-Q and GCC-Q. Graph-AC and Graph-Q are trained without considering cognitive-dissonance loss (CD-loss), so there is no guarantee of cognitive consistency. In contrast, GCC-AC and GCC-Q are regularized by the Global Cognitive Consistency. That is to say, all neighborhoods share the same hidden cognitive variable $C$, and thus all agents will form consistent cognitions on a global scale.

\subsection{Settings}
In our experiments, the weights of CD-loss are $\alpha=0.1$, $\alpha=0.1$ and $\alpha=0.2$ for packet routing, wifi configuration and Google football, respectively. In order to accelerate learning in Google football, we reduce the action set to \{top-right-move, bottom-right-move, high-pass, shot\} during exploration, and we give a reward $1.0$, $0.7$ and $0.3$ to the last three actions respectively if the agents score a goal. 

\subsection{Results}
\textbf{Results of Packet Routing.} The average rewards of 10 experiments are shown in Figure \ref{fig:PacketRoutingResults}. For the small topology, all methods achieve good performance. It means that the routing environments are suitable for testing RL methods.

For the middle topology (with 12 routers and 20 paths), ATT-MADDPG has better performance than MADDPG due to its advanced attention mechanism. Nevertheless, NCC-AC can obtain more rewards than the state-of-the-art ATT-MADDPG and MADDPG. It verifies the potential ability of our NCC-AC.

When the evaluation changes to the large topology (with 24 routers and 128 paths), ATT-MADDPG and MADDPG do not work at all, while NCC-AC can still maintain its performance. It indicates that NCC-AC has better scalability. The reason is that the interactions among agents become complicated and intractable when the number of agents becomes large. ATT-MADDPG and MADDPG simply treat all agents as a whole, which can hardly produce a good control policy. In contrast, NCC-AC decomposes all agents into much smaller neighborhoods, making agent interactions much simple and tractable. This property enables NCC-AC to work well even within a complex environment with an increasing number of agents.

\begin{figure}[!htb]
	\centering
	\begin{subfigure}{.25\textwidth}
		\includegraphics[width=.9\textwidth]{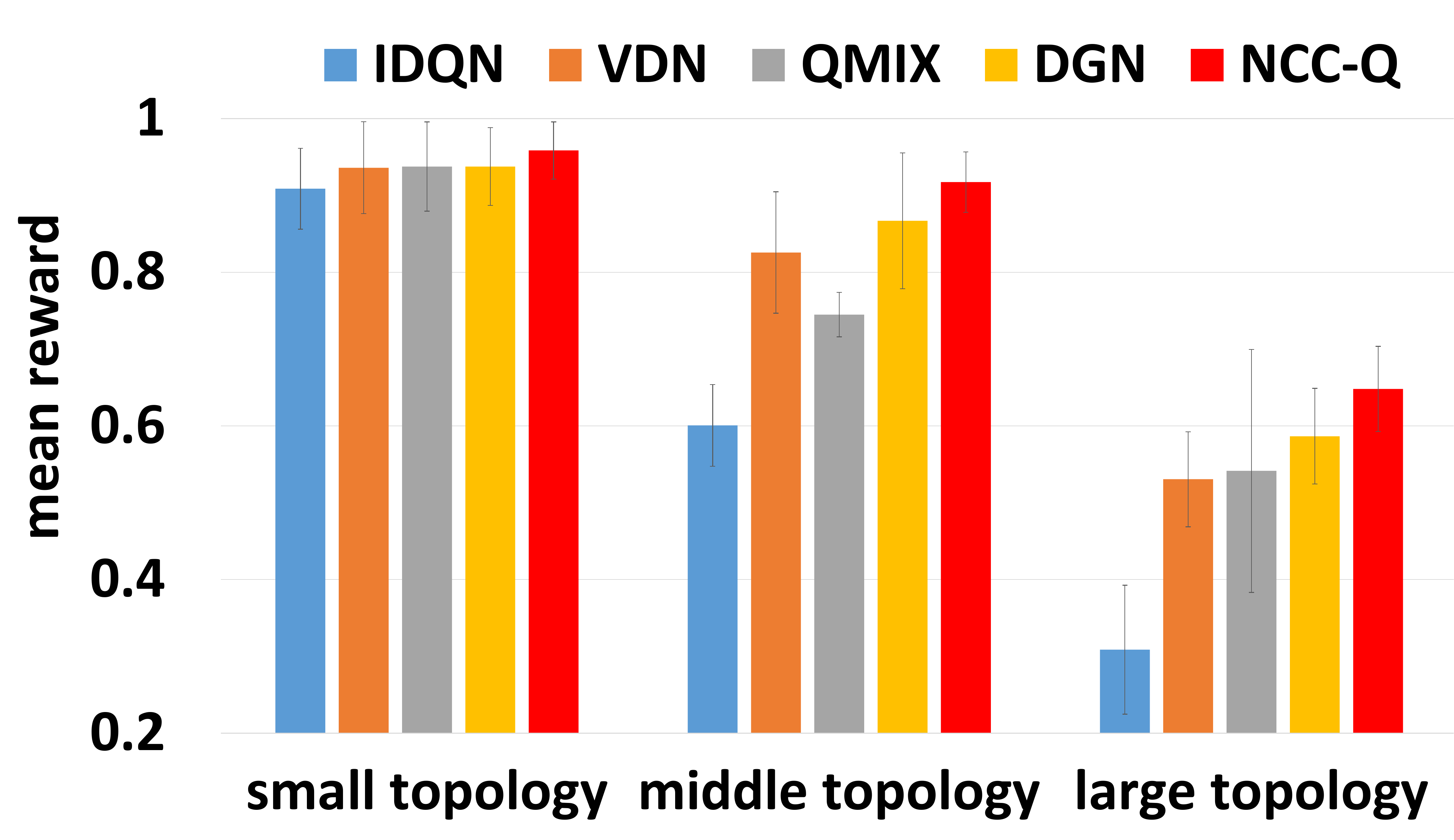}
		\caption{Wifi configuration tasks.}
		\label{fig:WifiResults}
	\end{subfigure}
	\begin{subfigure}{.18\textwidth}
		\includegraphics[width=.9\textwidth]{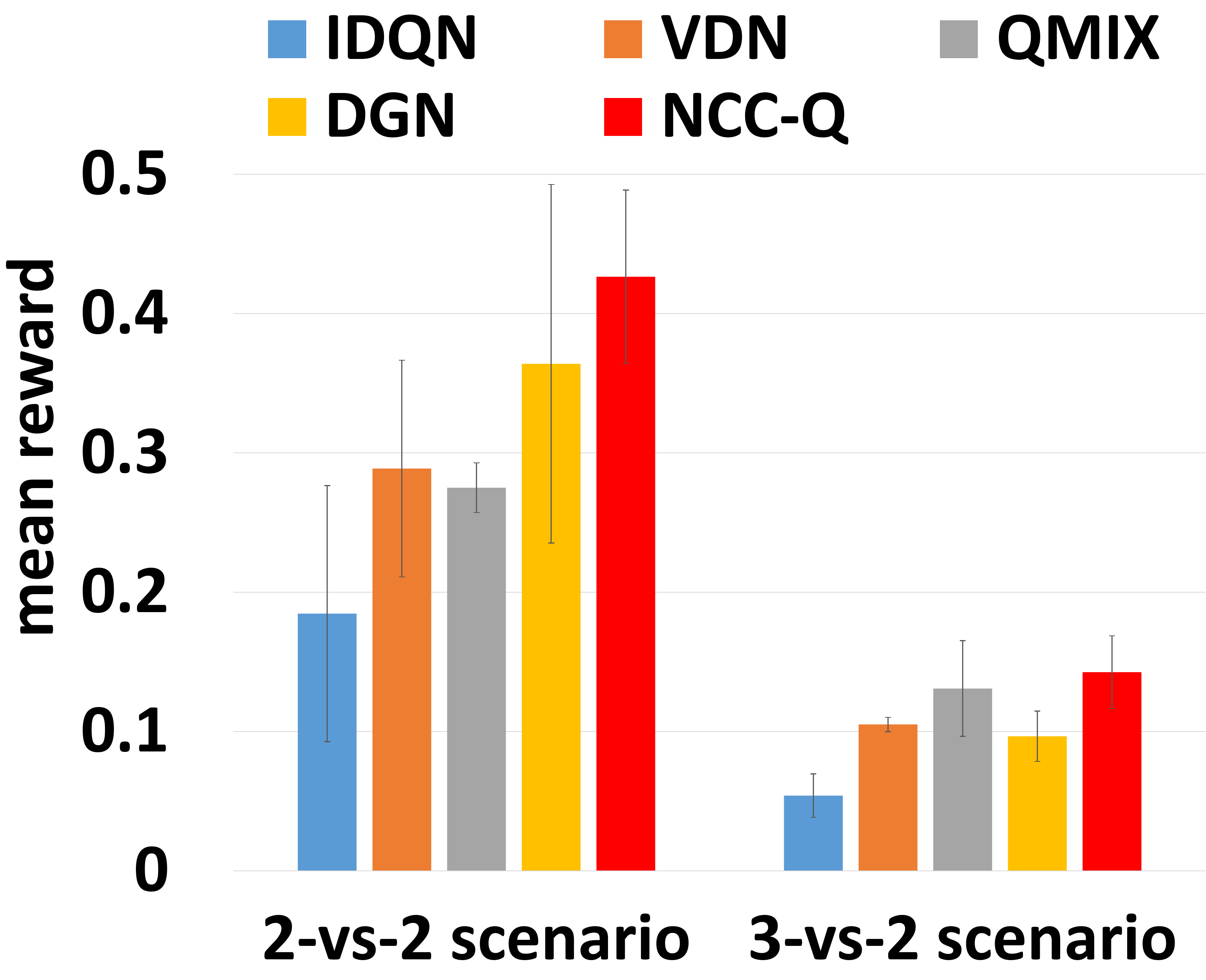}
		\caption{Google football tasks.}
		\label{fig:FootballResults}
	\end{subfigure}
	\caption{The average results of wifi and football tasks.}
	\label{fig:WifiFootballResults}
\end{figure}

\textbf{Results of Wifi and Football.} Figure \ref{fig:WifiResults} and \ref{fig:FootballResults} show the average rewards of 10 experiments for wifi configuration and Google football, respectively.

As can be seen, NCC-Q always obtains more rewards than all other methods in different topologies for both tasks. It indicates that our method is quite general and robust against different scenarios.

In addition, DGN achieves good mean rewards because the homogeneous environments are in favor of the parameter sharing design. However, DGN usually has a large variance. The comparison of VDN and QMIX shows that VDN outperforms QMIX in simple scenarios, while it underperforms QMIX in complex scenarios. This highlights the relationship among method performance, method complexity and task complexity. IDQN is always the worst one because it does not have any explicit coordination mechanism (like joint Q decomposition in VDN and QMIX, or agent decomposition in DGN and NCC-Q), which in turn shows the necessity of coordination mechanisms in large-scale settings.

\begin{figure}[!htb]
	\centering
	\begin{subfigure}{.23\textwidth}
		\includegraphics[width=.95\textwidth]{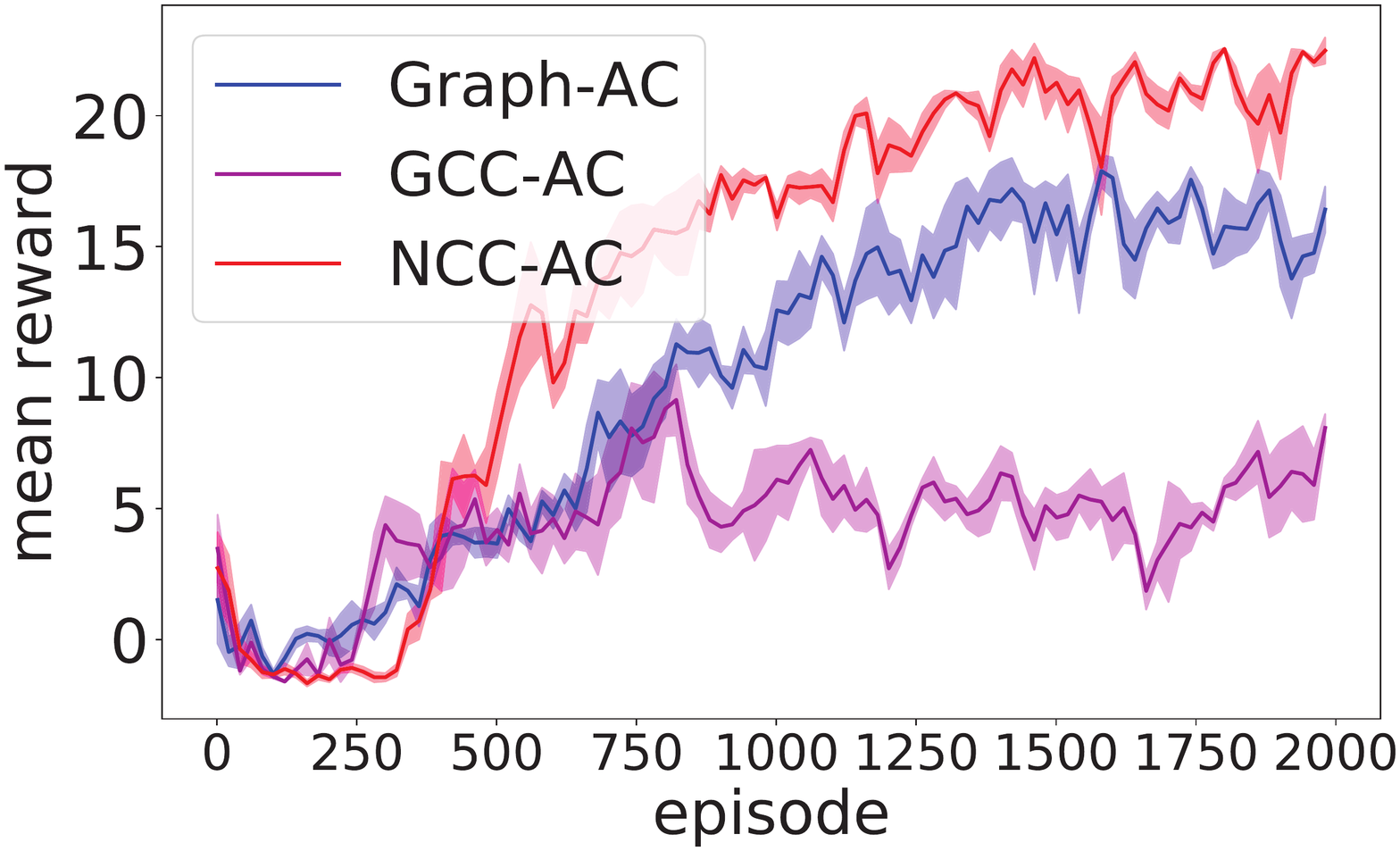}
		\caption{For middle topology.}
		\label{fig:large-Ablation}
	\end{subfigure}
	\begin{subfigure}{.23\textwidth}
		\includegraphics[width=.95\textwidth]{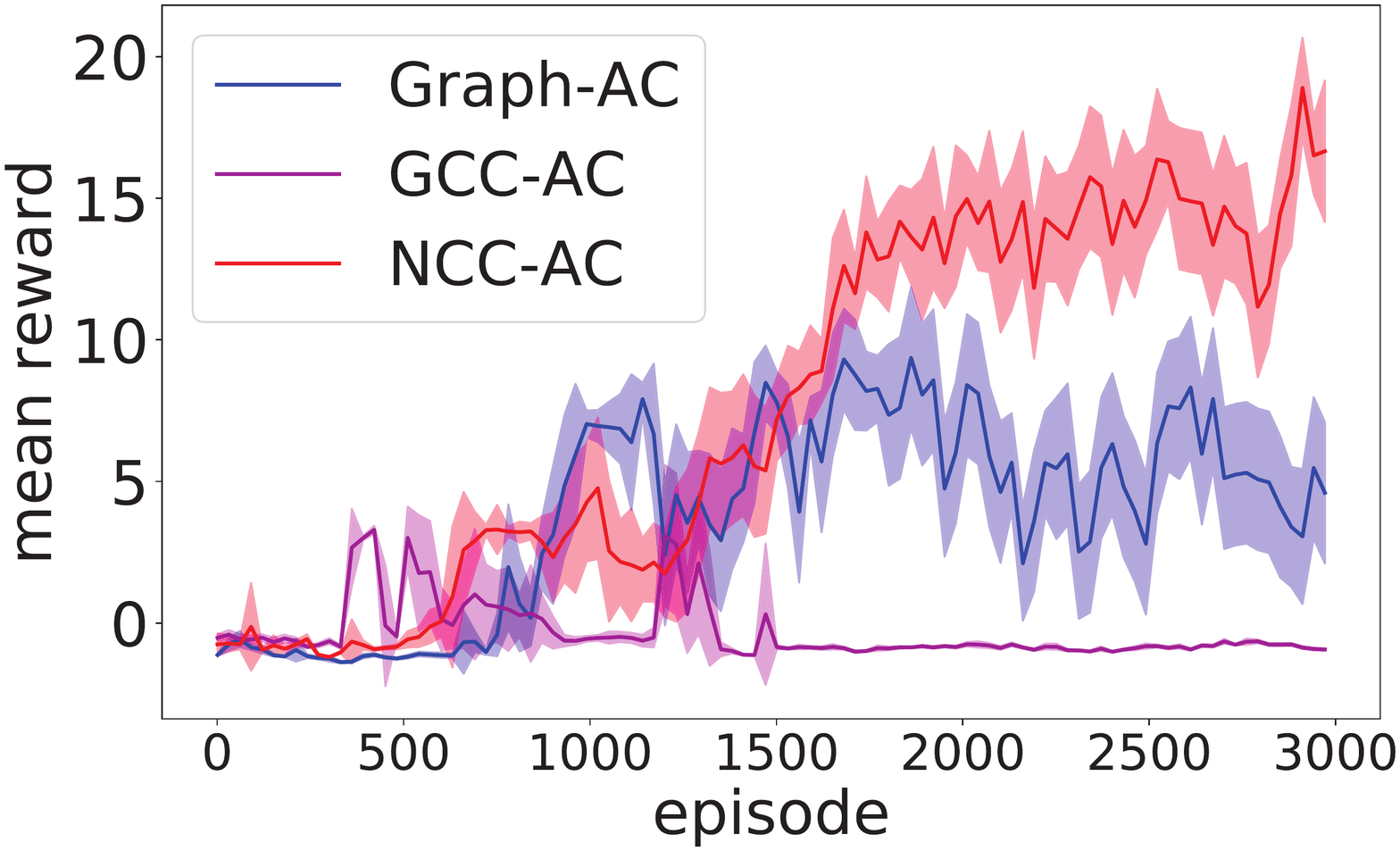}
		\caption{For large topology.}
		\label{fig:huge-Ablation}
	\end{subfigure}
	\caption{The ablation results of packet routing tasks.}
	\label{fig:PacketRoutingResults-Ablation}
\end{figure}

\textbf{Ablation Study on Cognitive Consistency.} The average results of packet routing tasks are shown in Figure \ref{fig:PacketRoutingResults-Ablation}. NCC-AC works better than Graph-AC in all topologies, while the performance of GCC-AC turns out to be unsatisfactory. It means that the neighborhood cognitive consistency (rather than global consistency or without consistency) really plays a critical role for achieving good results. Besides, GCC-AC works worse than Graph-AC in all topologies, which implies that it is harmful to enforce global cognitive consistency in the packet routing scenarios.

\begin{figure}[!htb]
	\centering
	\begin{subfigure}{.27\textwidth}
		\includegraphics[width=.95\textwidth]{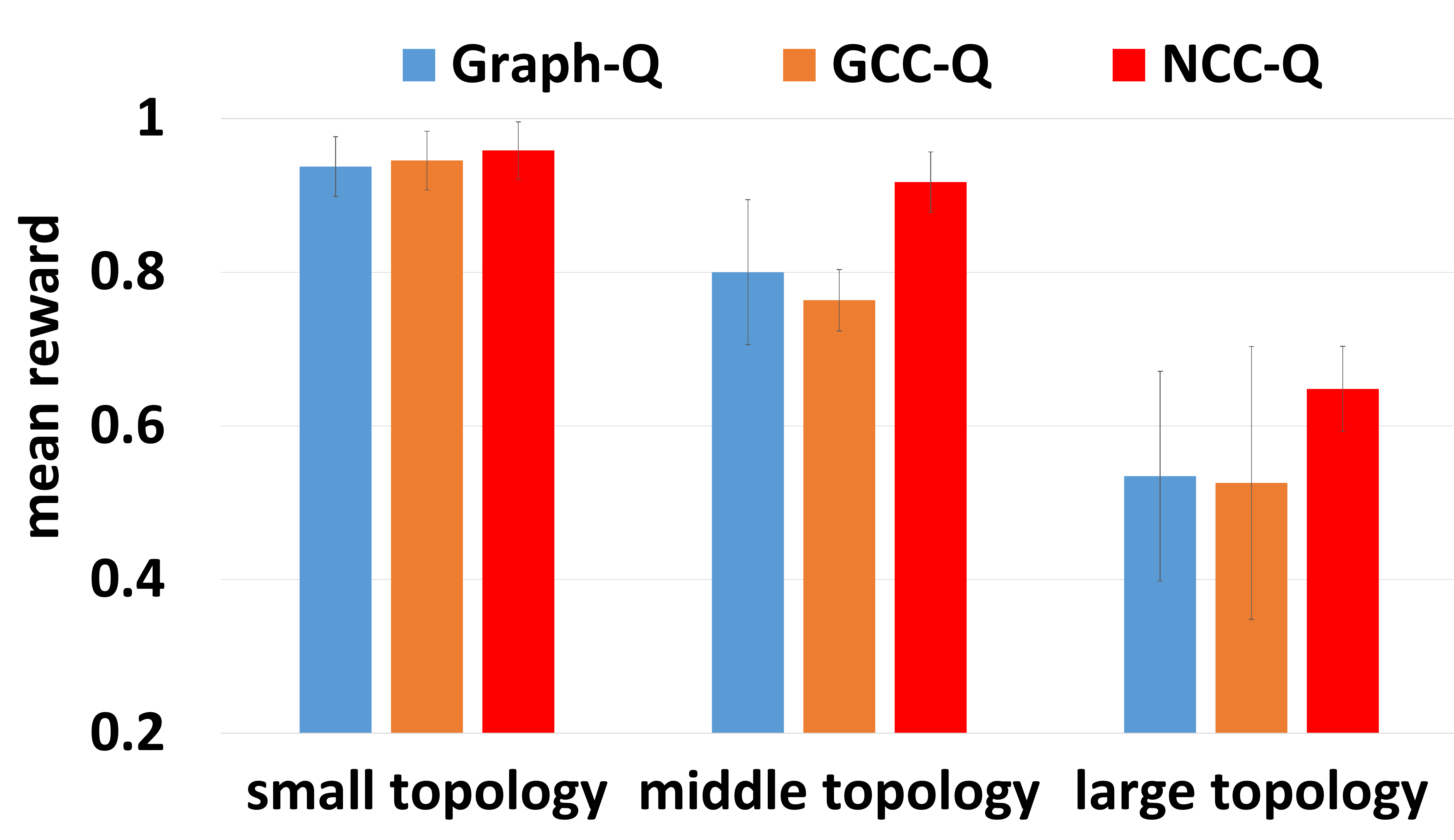}
		\caption{Wifi configuration tasks.}
		\label{fig:WifiResults-Ablation}
	\end{subfigure}
	\begin{subfigure}{.18\textwidth}
		\includegraphics[width=.95\textwidth]{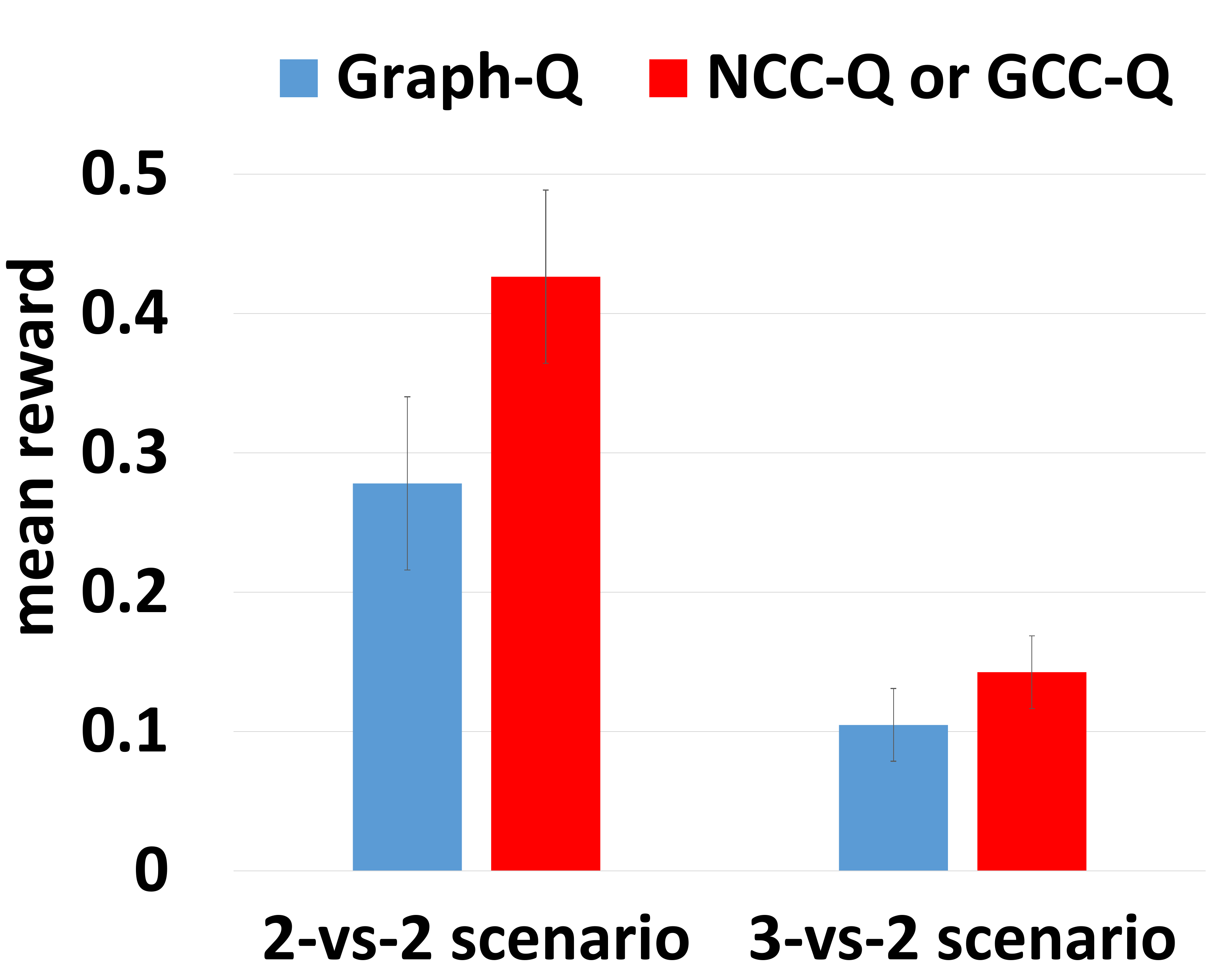}
		\caption{Google football tasks.}
		\label{fig:FootballResult-Ablations}
	\end{subfigure}
	\caption{The ablation results of wifi configuration and Google football tasks. For Google football, there is only one neighborhood, therefore GCC-Q is equivalent to NCC-Q.}
	\label{fig:WifiFootballResults-Ablation}
\end{figure}

The results of wifi and football tasks are shown in Figure \ref{fig:WifiFootballResults-Ablation}. NCC-Q has better performance than Graph-Q and GCC-Q, which is consistent with the results of routing scenarios. Thus, we draw the same conclusion that neighborhood cognitive consistency is the main reason for the good performance of NCC-Q. However, in contrast to the unsatisfactory results of GCC-AC in routing tasks, GCC-Q works pretty good (e.g., comparable with the state-of-the-art VDN and QMIX) in the wifi tasks. This is because the agents are homogeneous in wifi environments. They are more likely to form consistent cognitions on a global scale even \emph{without} ``CD-loss'', which makes GCC-Q a reasonable method. 

Overall, approaches with neighborhood cognitive consistency always work well in different scenarios, but methods with global cognitive consistency or without cognitive consistency can only achieve good results in specific tasks.

\textbf{Further Analysis on Cognitive Consistency.} 
We further verify how neighborhood cognitive consistency affects the learning process by training our methods with different loss settings: (1) only TD-loss, (2) the combination of TD-loss with CD-loss. The analyses are conducted based on the 2-vs-2 football scenario, because it is easy to understand by humans, and similar results can be found for other settings. We test two difficulty levels specified by game configurations \footnote{\url{https://github.com/google-research/football/blob/master/gfootball/env/config.py}}: ``game\_difficulty=0.6'' and ``game\_difficulty=0.9''.

\begin{figure}[!htb]
	\centering
	\begin{subfigure}{.23\textwidth}
		\includegraphics[width=.9\textwidth]{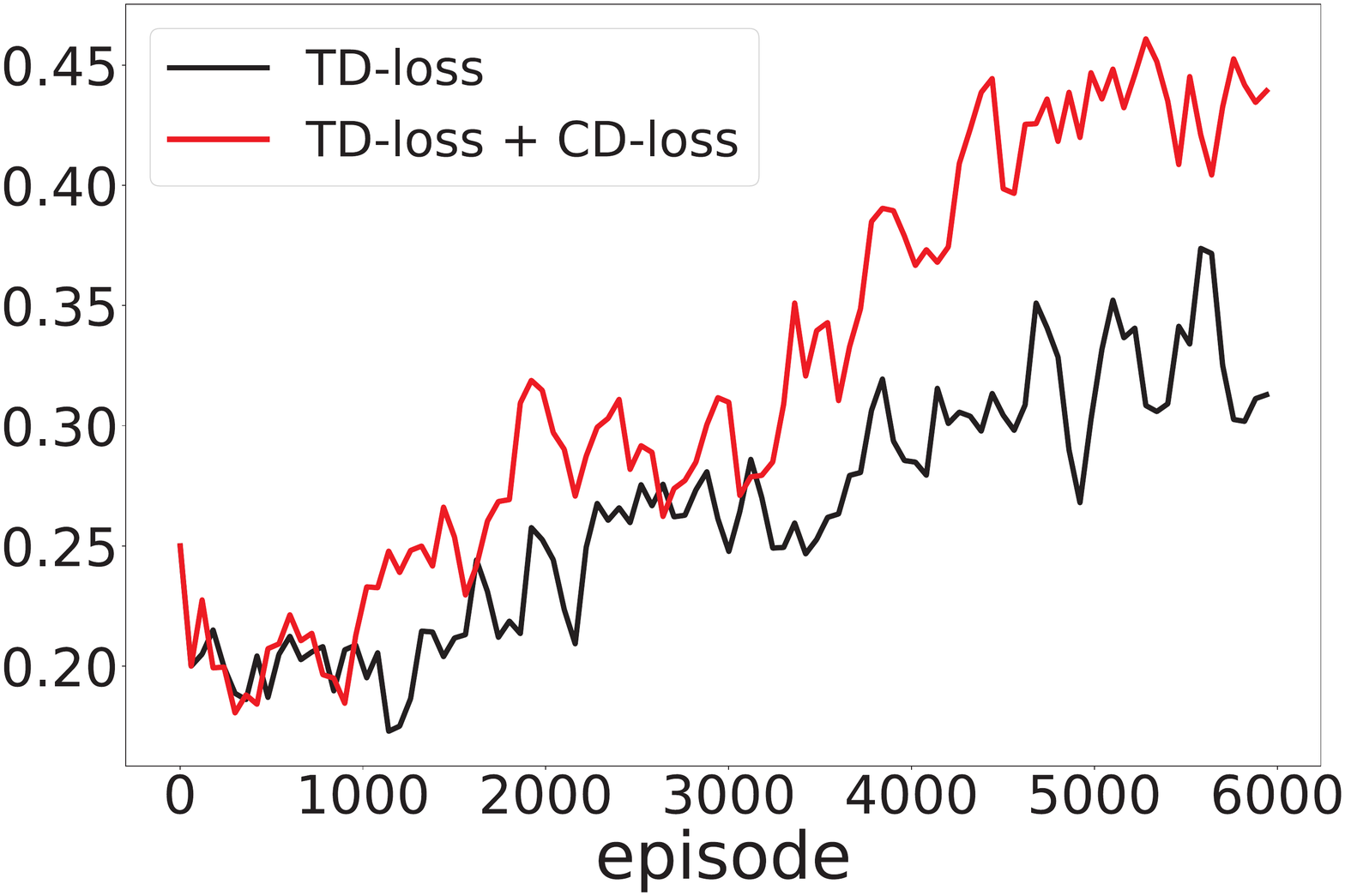}
		\caption{The mean reward.}
		\label{fig:CognitiveConsistencyAnalyses-mean_reward-football-zero6}
	\end{subfigure}
	\begin{subfigure}{.23\textwidth}
		\includegraphics[width=.9\textwidth]{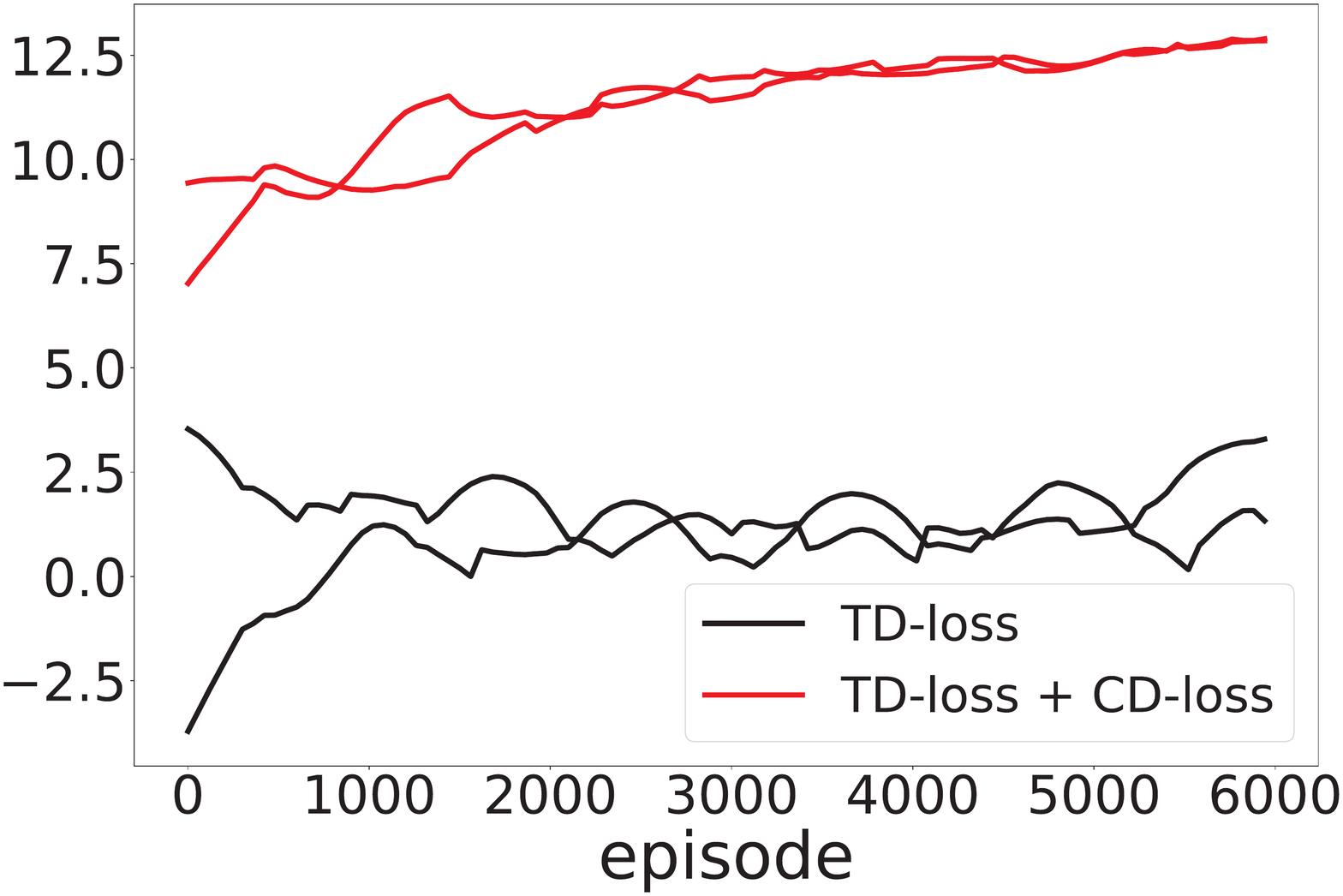}
		\caption{The cognition value.}
		\label{fig:CognitiveConsistencyAnalyses-baseline_value-football-zero6}
	\end{subfigure}
	\caption{The results of different loss settings for the 2-vs-2 football scenario with ``game\_difficulty=0.6''. In Figure (b), the cognition value stands for the arithmetic mean of all elements in variable $\widehat{C_i}$; besides, there are two curves belonging to two agents for each loss setting.}
	\label{fig:CognitiveConsistencyAnalyses-football-zero6}
\end{figure}

For ``game\_difficulty=0.6'', the learning curves are shown in Figure \ref{fig:CognitiveConsistencyAnalyses-football-zero6}. First, we can see that the mean rewards become greater and the cognition values of different agents become more consistent as the training goes on. Second, ``TD-loss + CD-loss'' achieves greater rewards and more consistent cognition values than a single ``TD-loss''. The results indicate that the proposed ``CD-loss'' plays a critical role to \emph{accelerate} the formation of cognitive consistency and thus better agent cooperations in the \emph{low-difficulty} scenario.

\begin{figure}[!htb]
	\centering
	\begin{subfigure}{.23\textwidth}
		\includegraphics[width=.9\textwidth]{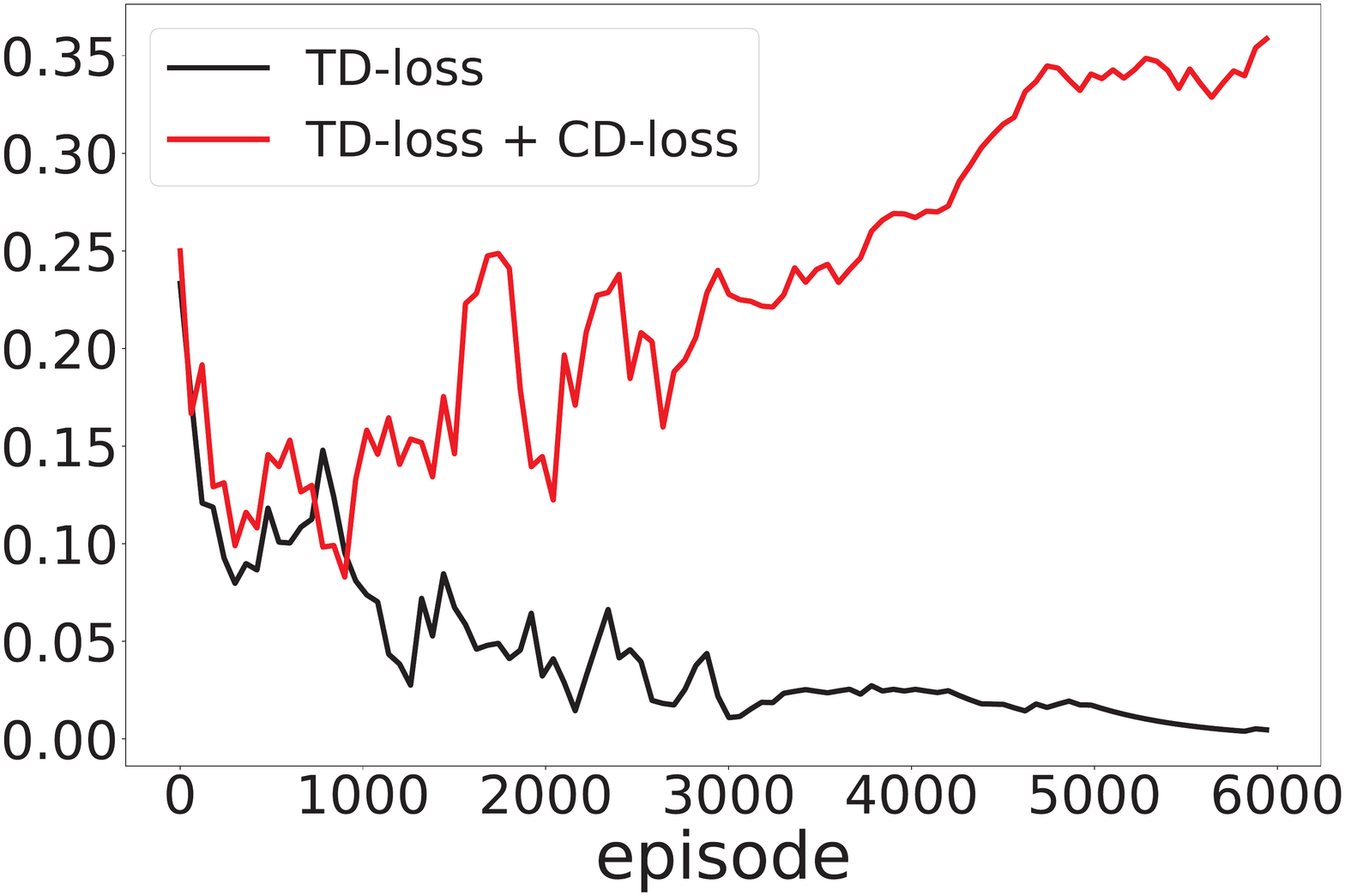}
		\caption{The mean reward.}
		\label{fig:CognitiveConsistencyAnalyses-mean_reward-football-zero9}
	\end{subfigure}
	\begin{subfigure}{.23\textwidth}
		\includegraphics[width=.9\textwidth]{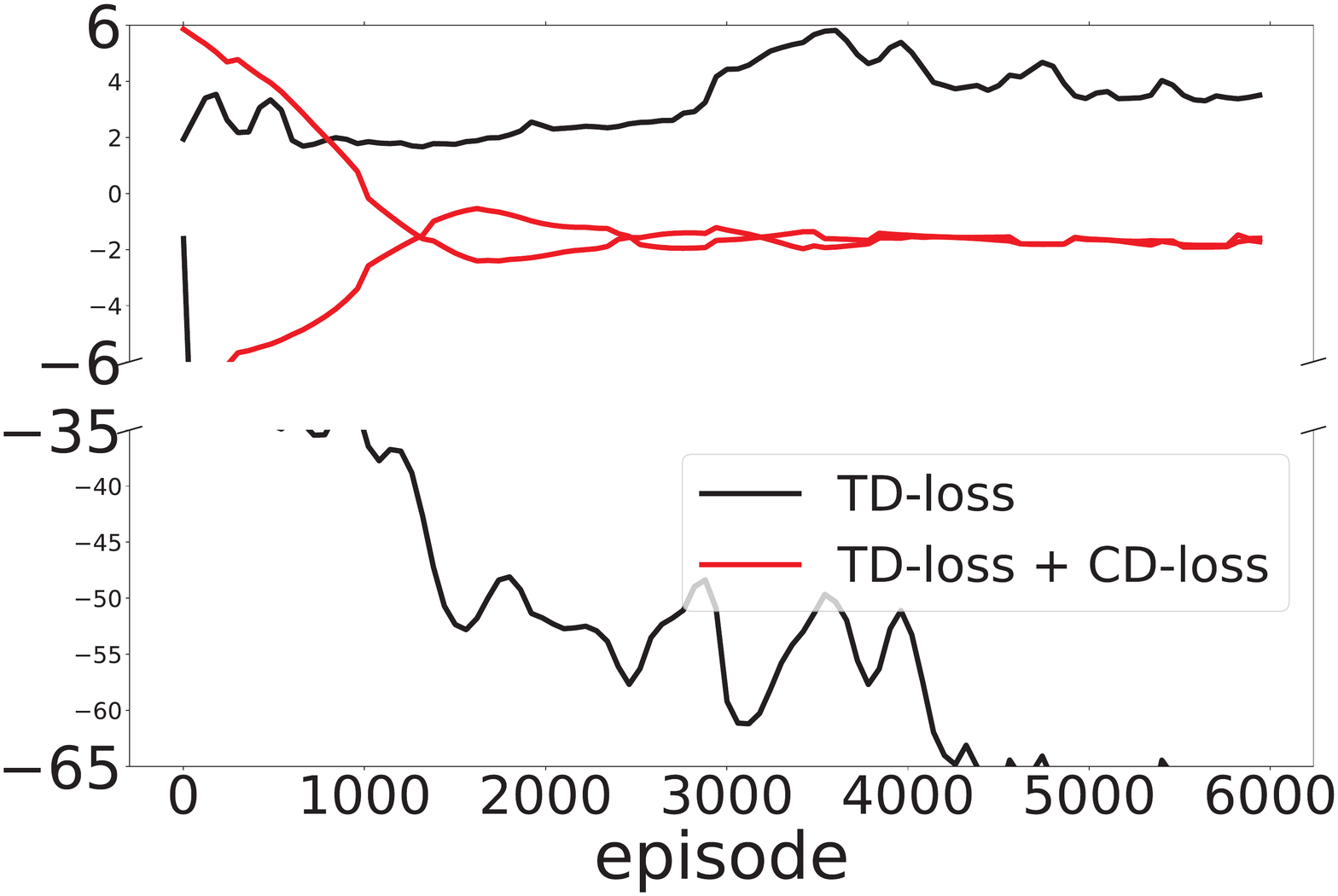}
		\caption{The cognition value.}
		\label{fig:CognitiveConsistencyAnalyses-baseline_value-football-zero9}
	\end{subfigure}
	\caption{The results of different loss settings for the 2-vs-2 football scenario with ``game\_difficulty=0.9''.}
	\label{fig:CognitiveConsistencyAnalyses-football-zero9}
\end{figure}

For ``game\_difficulty=0.9'', the learning curves are shown in Figure \ref{fig:CognitiveConsistencyAnalyses-football-zero9}. As can be observed, the mean rewards of ``TD-loss'' are very small, and the corresponding cognition values of the agents are very inconsistent. In contrast, the rewards of ``TD-loss + CD-loss'' are much greater than that of a single ``TD-loss''; correspondingly, ``TD-loss + CD-loss'' generates much more consistent cognition values for the agents. The results imply that cognitive consistency is critical for good performance in the \emph{high-difficulty} setting, where ``CD-loss'' has the ability to \emph{guarantee} the formation of cognitive consistency and thus better agent cooperations.

Considering all experiments, the most important lesson learned here is that there is usually a close relationship (e.g., positive correlation) between agent cooperation and agent cognitive consistency: if the agents have formed more similar and consistent cognition values, they are more likely to achieve better cooperations.

\section{Conclusion}
Inspired by both social psychology and real experiences, this paper introduces two novel neighborhood cognition consistent reinforcement learning methods, NCC-Q and NCC-AC, to facilitate large-scale agent cooperations. Our methods assume a hidden cognitive variable in each neighborhood, then infer this hidden cognitive variable by variational inference. As a result, all neighboring agents will eventually form consistent neighborhood cognitions and achieve good cooperations. We evaluate our methods on three tasks developed based on eight real-world scenarios. Extensive results show that they not only outperform the state-of-the-art methods by a clear margin, but also achieve good scalability in routing tasks. Moreover, ablation studies and further analyses are provided for better understanding of our methods.

\section{Acknowledgments}
The authors would like to thank Jun Qian, Dong Chen and Min Cheng for partial engineering support. The authors would also like to thank the anonymous reviewers for their comments. This work was supported by the National Natural Science Foundation of China under Grant No.61872397. The contact author is Zhen Xiao.

\bibliography{2313.aaai20} 

\begin{thebibliography}{}

\bibitem[\protect\citeauthoryear{Bear, Kagan, and Rand}{2017}]{bear2017co}
Bear, A.; Kagan, A.; and Rand, D.~G.
\newblock 2017.
\newblock Co-evolution of cooperation and cognition: the impact of imperfect
  deliberation and context-sensitive intuition.
\newblock {\em Proceedings of the Royal Society B: Biological Sciences}
  284(1851):20162326.

\bibitem[\protect\citeauthoryear{Bernstein \bgroup et al\mbox.\egroup
  }{2002}]{bernstein2002complexity}
Bernstein, D.~S.; Givan, R.; Immerman, N.; and Zilberstein, S.
\newblock 2002.
\newblock The complexity of decentralized control of markov decision processes.
\newblock {\em Mathematics of operations research} 27(4):819--840.

\bibitem[\protect\citeauthoryear{Blei, Kucukelbir, and
  McAuliffe}{2017}]{blei2017variational}
Blei, D.~M.; Kucukelbir, A.; and McAuliffe, J.~D.
\newblock 2017.
\newblock Variational inference: A review for statisticians.
\newblock {\em Journal of the American Statistical Association}
  112(518):859--877.

\bibitem[\protect\citeauthoryear{Corgnet, Esp{\'\i}n, and
  Hern{\'a}n-Gonz{\'a}lez}{2015}]{corgnet2015cognitive}
Corgnet, B.; Esp{\'\i}n, A.~M.; and Hern{\'a}n-Gonz{\'a}lez, R.
\newblock 2015.
\newblock The cognitive basis of social behavior: cognitive reflection
  overrides antisocial but not always prosocial motives.
\newblock {\em Frontiers in behavioral neuroscience} 9:287.

\bibitem[\protect\citeauthoryear{Duan \bgroup et al\mbox.\egroup
  }{2016}]{duan2016benchmarking}
Duan, Y.; Chen, X.; Houthooft, R.; Schulman, J.; and Abbeel, P.
\newblock 2016.
\newblock Benchmarking deep reinforcement learning for continuous control.
\newblock In {\em International Conference on Machine Learning},  1329--1338.

\bibitem[\protect\citeauthoryear{Foerster \bgroup et al\mbox.\egroup
  }{2016}]{foerster2016learning}
Foerster, J.; Assael, I.~A.; de~Freitas, N.; and Whiteson, S.
\newblock 2016.
\newblock Learning to communicate with deep multi-agent reinforcement learning.
\newblock In {\em Advances in Neural Information Processing Systems},
  2137--2145.

\bibitem[\protect\citeauthoryear{Foerster \bgroup et al\mbox.\egroup
  }{2018}]{foerster2018counterfactual}
Foerster, J.~N.; Farquhar, G.; Afouras, T.; Nardelli, N.; and Whiteson, S.
\newblock 2018.
\newblock Counterfactual multi-agent policy gradients.
\newblock In {\em Thirty-Second AAAI Conference on Artificial Intelligence}.

\bibitem[\protect\citeauthoryear{Jiang, Dun, and Lu}{2018}]{jiang2018graph}
Jiang, J.; Dun, C.; and Lu, Z.
\newblock 2018.
\newblock Graph convolutional reinforcement learning for multi-agent
  cooperation.
\newblock {\em arXiv preprint arXiv:1810.09202}.

\bibitem[\protect\citeauthoryear{Kurach \bgroup et al\mbox.\egroup
  }{2019}]{kurach2019google}
Kurach, K.; Raichuk, A.; Stanczyk, P.; Zajac, M.; Bachem, O.; Espeholt, L.;
  Riquelme, C.; Vincent, D.; Michalski, M.; Bousquet, O.; et~al.
\newblock 2019.
\newblock Google research football: A novel reinforcement learning environment.
\newblock {\em arXiv preprint arXiv:1907.11180}.

\bibitem[\protect\citeauthoryear{Lakkaraju and
  Speed}{2019}]{lakkaraju2019cognitive}
Lakkaraju, K., and Speed, A.
\newblock 2019.
\newblock A cognitive-consistency based model of population wide attitude
  change.
\newblock In {\em Complex Adaptive Systems}. Springer.
\newblock  17--38.

\bibitem[\protect\citeauthoryear{Lowe \bgroup et al\mbox.\egroup
  }{2017}]{Lowe2017Multi}
Lowe, R.; Wu, Y.; Tamar, A.; Harb, J.; Abbeel, O.~P.; and Mordatch, I.
\newblock 2017.
\newblock Multi-agent actor-critic for mixed cooperative-competitive
  environments.
\newblock In {\em Advances in Neural Information Processing Systems},
  6379--6390.

\bibitem[\protect\citeauthoryear{Mao \bgroup et al\mbox.\egroup
  }{2019}]{mao2019modelling}
Mao, H.; Zhang, Z.; Xiao, Z.; and Gong, Z.
\newblock 2019.
\newblock Modelling the dynamic joint policy of teammates with attention
  multi-agent {DDPG}.
\newblock In {\em Proceedings of the 18th International Conference on
  Autonomous Agents and MultiAgent Systems},  1108--1116.
\newblock International Foundation for Autonomous Agents and Multiagent
  Systems.

\bibitem[\protect\citeauthoryear{Mnih \bgroup et al\mbox.\egroup
  }{2015}]{mnih2015human}
Mnih, V.; Kavukcuoglu, K.; Silver, D.; Rusu, A.~A.; Veness, J.; Bellemare,
  M.~G.; Graves, A.; Riedmiller, M.; Fidjeland, A.~K.; Ostrovski, G.; et~al.
\newblock 2015.
\newblock Human-level control through deep reinforcement learning.
\newblock {\em Nature} 518(7540):529.

\bibitem[\protect\citeauthoryear{OroojlooyJadid and
  Hajinezhad}{2019}]{oroojlooyjadid2019review}
OroojlooyJadid, A., and Hajinezhad, D.
\newblock 2019.
\newblock A review of cooperative multi-agent deep reinforcement learning.
\newblock {\em arXiv preprint arXiv:1908.03963}.

\bibitem[\protect\citeauthoryear{Peng \bgroup et al\mbox.\egroup
  }{2017}]{peng2017multiagent}
Peng, P.; Yuan, Q.; Wen, Y.; Yang, Y.; Tang, Z.; Long, H.; and Wang, J.
\newblock 2017.
\newblock Multiagent bidirectionally-coordinated nets for learning to play
  starcraft combat games.
\newblock {\em arXiv preprint arXiv:1703.10069}.

\bibitem[\protect\citeauthoryear{Rashid \bgroup et al\mbox.\egroup
  }{2018}]{rashid2018qmix}
Rashid, T.; Samvelyan, M.; Witt, C.~S.; Farquhar, G.; Foerster, J.; and
  Whiteson, S.
\newblock 2018.
\newblock Qmix: Monotonic value function factorisation for deep multi-agent
  reinforcement learning.
\newblock In {\em International Conference on Machine Learning},  4292--4301.

\bibitem[\protect\citeauthoryear{Russo \bgroup et al\mbox.\egroup
  }{2008}]{russo2008goal}
Russo, J.~E.; Carlson, K.~A.; Meloy, M.~G.; and Yong, K.
\newblock 2008.
\newblock The goal of consistency as a cause of information distortion.
\newblock {\em Journal of Experimental Psychology: General} 137(3):456.

\bibitem[\protect\citeauthoryear{Silver \bgroup et al\mbox.\egroup
  }{2014}]{silver2014deterministic}
Silver, D.; Lever, G.; Heess, N.; Degris, T.; Wierstra, D.; and Riedmiller, M.
\newblock 2014.
\newblock Deterministic policy gradient algorithms.
\newblock In {\em International Conference on Machine Learning},  387--395.

\bibitem[\protect\citeauthoryear{Simon, Snow, and Read}{2004}]{simon2004redux}
Simon, D.; Snow, C.~J.; and Read, S.~J.
\newblock 2004.
\newblock The redux of cognitive consistency theories: evidence judgments by
  constraint satisfaction.
\newblock {\em Journal of personality and social psychology} 86(6):814.

\bibitem[\protect\citeauthoryear{Son \bgroup et al\mbox.\egroup
  }{2019}]{son2019qtran}
Son, K.; Kim, D.; Kang, W.~J.; Hostallero, D.~E.; and Yi, Y.
\newblock 2019.
\newblock Qtran: Learning to factorize with transformation for cooperative
  multi-agent reinforcement learning.
\newblock In {\em International Conference on Machine Learning},  5887--5896.

\bibitem[\protect\citeauthoryear{Sukhbaatar, Fergus, and
  others}{2016}]{sukhbaatar2016learning}
Sukhbaatar, S.; Fergus, R.; et~al.
\newblock 2016.
\newblock Learning multiagent communication with backpropagation.
\newblock In {\em Advances in Neural Information Processing Systems},
  2244--2252.

\bibitem[\protect\citeauthoryear{Sunehag \bgroup et al\mbox.\egroup
  }{2018}]{sunehag2018value}
Sunehag, P.; Lever, G.; Gruslys, A.; Czarnecki, W.~M.; Zambaldi, V.; Jaderberg,
  M.; Lanctot, M.; Sonnerat, N.; Leibo, J.~Z.; Tuyls, K.; et~al.
\newblock 2018.
\newblock Value-decomposition networks for cooperative multi-agent learning
  based on team reward.
\newblock In {\em Proceedings of the 17th International Conference on
  Autonomous Agents and MultiAgent Systems},  2085--2087.
\newblock International Foundation for Autonomous Agents and Multiagent
  Systems.

\bibitem[\protect\citeauthoryear{Sutton and
  Barto}{1998}]{sutton1998introduction}
Sutton, R.~S., and Barto, A.~G.
\newblock 1998.
\newblock {\em Introduction to reinforcement learning}, volume~2.
\newblock MIT press Cambridge.

\bibitem[\protect\citeauthoryear{Tampuu \bgroup et al\mbox.\egroup
  }{2017}]{tampuu2017multiagent}
Tampuu, A.; Matiisen, T.; Kodelja, D.; Kuzovkin, I.; Korjus, K.; Aru, J.; Aru,
  J.; and Vicente, R.
\newblock 2017.
\newblock Multiagent cooperation and competition with deep reinforcement
  learning.
\newblock {\em PloS one} 12(4):e0172395.

\bibitem[\protect\citeauthoryear{Tan}{1993}]{tan1993multi}
Tan, M.
\newblock 1993.
\newblock Multi-agent reinforcement learning: Independent vs. cooperative
  agents.
\newblock In {\em Proceedings of the tenth international conference on machine
  learning},  330--337.

\end{thebibliography}
\bibliographystyle{aaai} 

\end{document}